\documentclass[10pt,twocolumn,letterpaper]{article}
\usepackage{multicol} 
\usepackage{colortbl}  
\usepackage{multirow}
\usepackage{color}  
\usepackage{pifont}

\definecolor{Graya}{gray}{0.9} 
 \usepackage[pagenumbers]{cvpr} 

\usepackage[dvipsnames]{xcolor}

\definecolor{cvprblue}{rgb}{0.21,0.49,0.74}
\usepackage[pagebackref,breaklinks,colorlinks,citecolor=cvprblue]{hyperref}

\def\paperID{10870} 
\def\confName{CVPR}
\def\confYear{2024}

\newcommand{\dec}[1]{\textcolor{blue}{#1}}
\newcommand{\inc}[1]{\textcolor{red}{#1}}
\newcommand\blfootnote[1]{%
  \begingroup
  \renewcommand\thefootnote{}\footnote{#1}%
  \addtocounter{footnote}{-1}%
  \endgroup
}

\title{LLaMA-Excitor: General Instruction Tuning via Indirect Feature Interaction}
\author{Bo Zou *\\
Tsinghua University\\
Beijing, China\\
{\tt\small zoub21@mails.tsinghua.edu.cn}
\and
Chao Yang *\\
Shanghai AI Laboratory\\
Shanghai, China\\
{\tt\small yangchao@pjlab.org.cn}
\and
Yu Qiao\\
Shanghai AI Laboratory\\
Shanghai, China\\
{\tt\small qiaoyu@pjlab.org.cn}
\and
Chengbin Quan\\
Tsinghua University\\
Beijing, China\\
{\tt\small quancb@tsinghua.edu.cn}
\and
Youjian Zhao \dag\\
Tsinghua University\\
Zhongguancun Laboratory\\
Beijing, China\\
{\tt\small zhaoyoujian@tsinghua.edu.cn}
}

\begin{document}
\maketitle
\blfootnote{
$^*$ Equal contribution, $^\dag$ Corresponding author \\
\indent\indent
This work was done during an internship at Shanghai AI Lab.
}


\vspace{-0.15cm}

\begin{abstract}
Existing methods to fine-tune LLMs, like Adapter, Prefix-tuning, and LoRA, which introduce extra modules or additional input sequences to inject new skills or knowledge, may compromise the innate abilities of LLMs.
In this paper, we propose LLaMA-Excitor, a lightweight method that stimulates the LLMs' potential to better follow instructions by gradually paying more attention to worthwhile information.
Specifically, LLaMA-Excitor does not directly change the intermediate hidden state during the self-attention calculation.
We designed the Excitor block as a bypass module that reconstructs Keys and changes the importance of Values in self-attention using learnable prompts. LLaMA-Excitor ensures a self-adaptive allocation of additional attention to input instructions, thus effectively preserving LLMs' pre-trained knowledge when fine-tuning LLMs on low-quality instruction-following datasets. 
Furthermore, we unify the modeling of multi-modal and language-only tuning, extending LLaMA-Excitor to a powerful visual instruction follower without the need for complex multi-modal alignment. 
Our approach is evaluated in language-only and multi-modal scenarios. Compared with the original LLaMA-7B, LLaMA-Excitor is the only PEFT method that maintains basic capabilities and achieves +3.12\% relative improvement on the MMLU benchmark.
In the visual instruction tuning, we achieve a new state-of-the-art image captioning performance on MSCOCO (157.5 CIDEr), and a comparable performance on ScienceQA (88.39\%) to cutting-edge models with more parameters and extensive vision-language pertaining. The code will be available at~\url{https://zoubo9034.github.io/Excitor/}.
\end{abstract}    
\section{Introduction}
\label{sec:intro}
\begin{figure}[!t]
    \centering
    \includegraphics[width=0.45\textwidth]{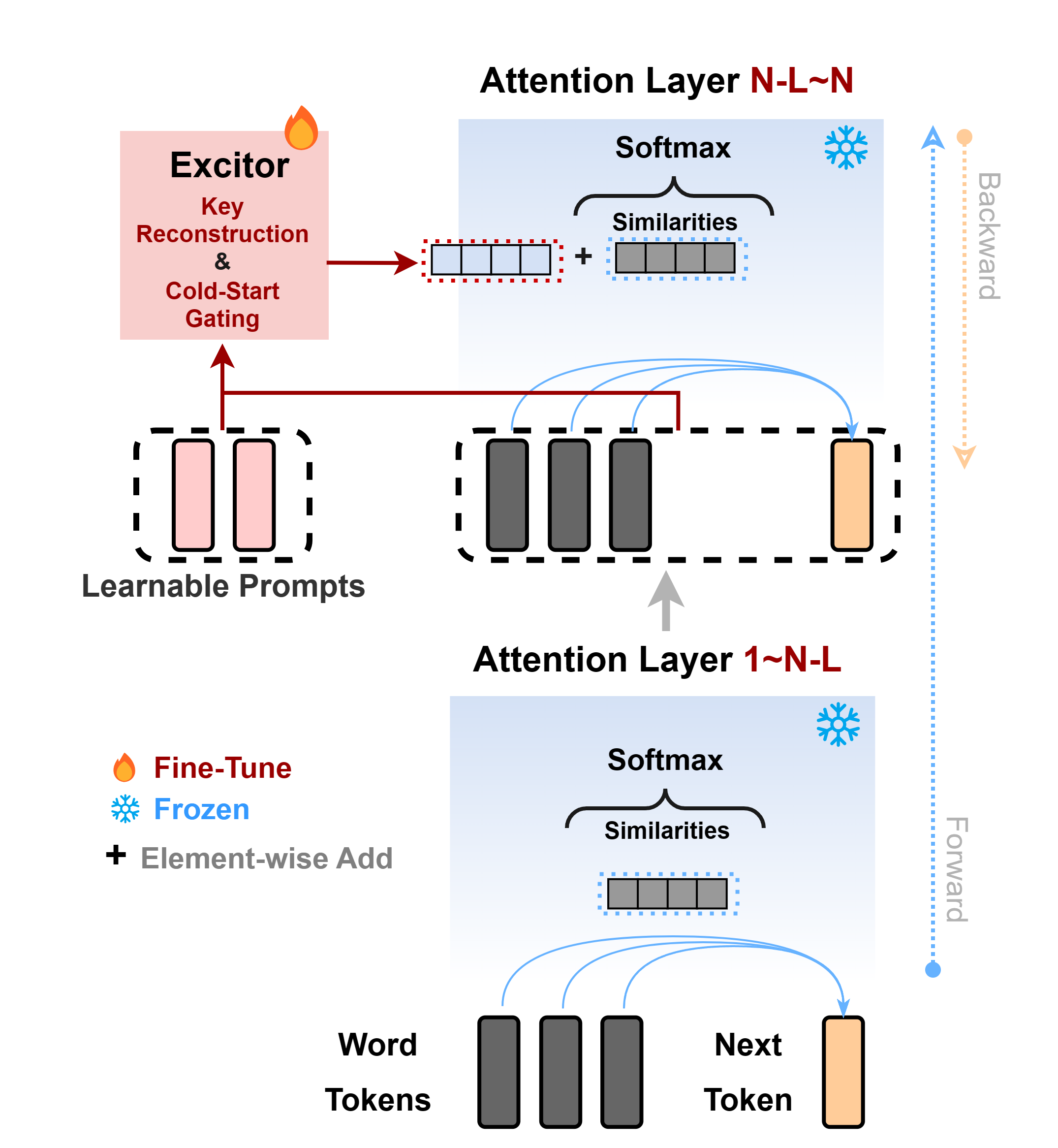}
    \caption{Overview of LLaMA-Excitor. We integrate Excitor blocks into L out of N attention layers of LLaMA. Differing from previous PEFT techniques, LLaMA-Excitor indirectly involves learnable information in the reasoning process by changing the similarity matrices. It ensures that the hidden states are within the original distribution of LLaMA.}
    \label{fig:Overview}
\end{figure}

Large language models (LLMs) are highly advanced tools that enable contextual information processing, understanding, and generation. They serve as powerful knowledge bases that can be used to gain valuable insights and create new content \cite{dai2019transformer,devlin2018bert,radford2019language,raffel2020exploring,zhang2022opt}. 
Fine-tuning them for downstream tasks has also achieved great success in various domains. With the rapid growth in scale and quality of the corpus and the advance in hardware, the inherent abilities of LLMs are becoming stronger. Therefore, the primary objective of fine-tuning general-purpose LLMs should be ensuring they produce the desired output for downstream tasks (i.e., instruction-following ability) rather than memory excessive knowledge or mastering skills (e.g., causal inference and numerical computation).

Prioritizing the enhancement of instruction-following capabilities for LLMs is prudent for a multitude of reasons. Firstly, it is often presumed that the corpus utilized during training encompasses a comprehensive repository of knowledge, thereby obviating the need for additional fine-tuning datasets when addressing specific downstream tasks. This implies that enriching a model's ability to follow instructions could leverage this pre-existing knowledge base more effectively. 
Secondly, by standardizing the output format during instruction-following tasks, we can reduce the generation of harmful or irrelevant content.
Third, the challenge of incorporating new knowledge and skills into LLMs is accentuated by the limited set of parameters available for fine-tuning. Lastly, fine-tuning in the context of multi-modal tasks can be succinctly expressed as enhancing a model's proficiency in responding to visual cues. This essentially constitutes a variant of instruction following, where the prompts are visual rather than textual.


However, predominant parameter-efficient fine-tuning (PEFT) techniques are met with notable challenges in performing instruction-tuning.
The Adapter methods \cite{houlsby2019parameter, pfeiffer2020adapterfusion}, which have a serial structure, can cause computation bottlenecks, and they also significantly alter the inference process to adapt to downstream tasks, which in turn can lead to degradation of inherent abilities, such as catastrophic forgetting. Prompt learning methods \cite{li2021prefix,liu2021p,brown2020language,schick2020exploiting} effectively incorporate new information by creating additional parameters corresponding to the input sequences of language models. However, concentrating all necessary knowledge into fixed-length token sequences is challenging for various downstream tasks. Furthermore, as the length of the generated sequence increases, the control over the sequence diminishes since LLMs generate the next word based on a softmax over the entire sequence. LoRA methods \cite{hu2021lora,zhang2023adaptive} directly add the outcomes of randomly initialized low-rank modules with linear layers' outputs, which can introduce features that are out of LLMs' feature distribution and may cause degradations. Recent studies \cite{zhang2023adaptive,hu2023vl} also show that Adapter and LoRA evenly assign learnable parameters to trainable modules and neglect the functionality gap between each layer in LLM, which will lead to performance drops.

In this paper, we start from a new perspective and propose LLaMA-Excitor, a PEFT method that focuses on the following instructions. (1) LLaMA-Excitor aims to optimize instruction-following ability by releasing the potential of an LLM, i.e., LLaMA \cite{touvron2302llama}, instead of pursuing new knowledge and skills. Due to variations in the data scale, quality, and content coverage of instruction-tuning sets, the effect of fine-tuning processes on the inherent skills of LLMs is unpredictable. (2) LLaMA-Excitor can reduce the degradation when fine-tuning on unsuited datasets. Specifically, rather than directly changing the intermediate hidden state of pretrained LLMs like Adapter and LoRA, LLaMA-Excitor uses trainable bypass modules (Excitor blocks) to modify the attention score meanwhile maintain original $Values$ for each attention layer in LLaMA. Excitor blocks only adjust the proportion of information in $Values$ for hidden states, ensuring that each attention layer's input is derived from a linear combination of the previous layer's $Values$ and corresponds to the original LLaMA's distribution (an indirect feature interaction process). Additionally, the modification of the attention score depends on both the input sequence and a set of prompt tokens to make the model adaptively decide whether to rely on additional information from the fine-tuning set for attention allocation. These attributes ensure fast training and few forgetings.

Moreover, Exictor innovatively provides a low-budget way to fine-tune a language-only model into a vision-language model. Previous works (e.g., \cite{tsimpoukelli2021multimodal, mokady2021clipcap, zhang2023llama, liu2023visual, anonymous2024videodistill}) rely on training an additional visual branch or projection module to align vision and language, which incurs significant computational overhead. We explore the unified modeling of instruction-tuning and visual instruction-tuning for LLMs and utilize a visual Exictor to make LLaMA follow the visual prompts generated from a frozen image encoder like CLIP \cite{radford2021learning}. In this way, we eliminate the need for extra vision-language alignment while preserving textual features' purity during LLMs' reasoning and shortening the training time. Our contributions are summarized as follows: 
\begin{itemize}
\item We study indirect feature interaction in fine-tuning LLMs and propose LLaMA-Excitor, a PEFT method that focuses on instruction-following and reduces forgetting.
\item We uniformly models multi-modal and language-only tuning and extends language models into powerful vision-language models in a low-budget way.
\item Experiments on instruction-following datasets, multi-task evaluation, image captioning, and VQA demonstrate the feasibility and advancement of LLaMA-Excitor.
\end{itemize}
\section{Related Works}

\begin{figure*}[!t]
    \centering
    \includegraphics[width=0.8\textwidth]{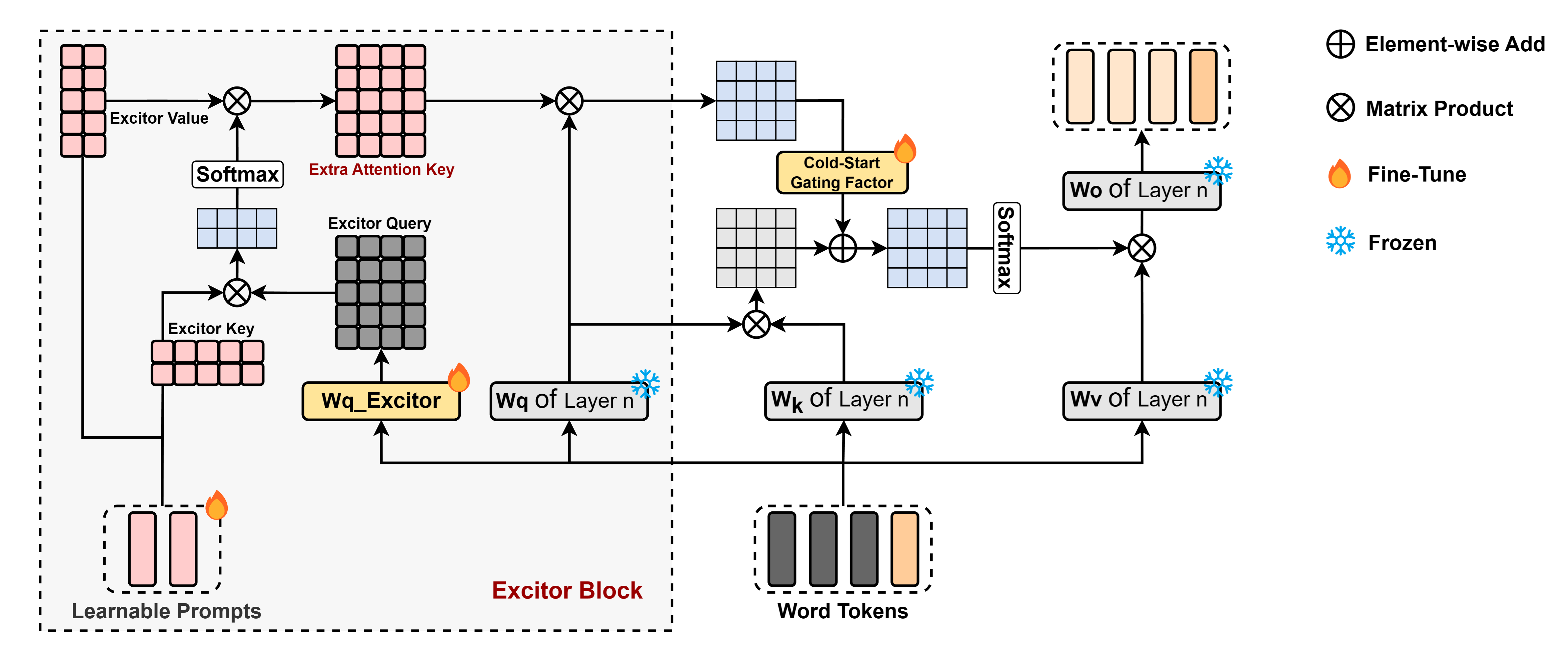}
    \caption{Details of the Excitor block. We assign a set of learnable prompts for attention layers of LLaMA. These prompts are used to construct an extra $Key$ for computing additional similarity scores, which are then merged into the original scores to alter the LLM's behavior. Cold-start gating factors are designed to stabilize the training.}
    \label{fig:Block Details}
     \vspace{-0.5cm}
\end{figure*}

\subsection{Parameter-Efficient Fine-Tuning}
PEFT techniques enable efficient adaptation of large language models to various downstream applications without fine-tuning all the model's parameters.
Adapter-based approaches \cite{houlsby2019parameter, pfeiffer2020adapterfusion} insert additional feed-forward networks within the existing architecture. However, these methods can increase the computational load due to their serial nature and may induce catastrophic forgetting by significantly changing the reasoning mechanism of LLMs.
Prompt-learning methods offer a training-free approach to task adaptation \cite{brown2020language,schick2020exploiting} and are improved by prefix-tuning \cite{li2021prefix,liu2021p}, which involves learnable token sequences. The primary drawback is the difficulty in capturing complex knowledge within a limited number of tokens, which can be restrictive for diverse tasks. Additionally, as the number of learnable tokens grows, these methods may lose control over the sequence output quality.
LoRA \cite{hu2021lora,zhang2023adaptive} aims to address these issues by introducing low-rank matrices in a parallel structure, enhancing both performance and efficiency. However, it risks incorporating features that deviate from the original model's distribution, potentially leading to performance degradation. Furthermore, equal distribution of parameter updates across layers disregards the distinct functionalities of different layers \cite{zhang2023adaptive,hu2023vl}, leading to suboptimal utilization of the model's capacity.

\subsection{Visual Instruction Tuning}
Instruction tuning \cite{ouyang2022training, wang2022self} finetunes or adapted language models to follow specific textual instructions.  Despite the prominent success in the NLP domain, its inherent text-only nature limits its ability to comprehend and interact with visual content effectively. LLaVA \cite{liu2023visual} enables language models to follow visual instructions and engage with
multi-modal information. And a series of approaches have been developed to improve the visual
instruction tuning with advanced architectures \cite{bai2023qwen, li2023blip, alayrac2022flamingo} or more versatile functionalities \cite{lai2023lisa, peng2023kosmos, zhu2023minigpt}. 
Fine-tuning in the context of multi-modal tasks can be succinctly expressed as enhancing a model's proficiency in responding to visual cues. This essentially constitutes a variant of instruction following, where the prompts are visual rather than textual.

\section{Methodology}
\label{sec:Method}
The primary goal of Excitor is to effectively leverage LLMs' pre-trained knowledge and skills to solve downstream tasks without excessive modification of the reasoning process, which can lead to forgetting and performance degradation. We choose LLaMA\cite{touvron2023llama} as our LLM as its effectiveness has been demonstrated in several open-source instruction-tuning works \cite{zhang2023llama,taori2023stanford,chiang2023vicuna,peng2023instruction,liu2023visual}. In Section \ref{indirect}, we first introduce how to insert Excitor blocks into LLaMA's attention layers. Then, in Section \ref{Excitor}, we present the details of Excitor blocks' architecture. Finally, in section \ref{Multi}, we generalize LLaMA-Excitor for multi-modal reasoning.

\subsection{Indirect feature interaction}
\label{indirect}
Given 52K instruction-following data \cite{taori2023stanford} and a pre-trained LLaMA with $N$ attention layers, we insert our proposed Excitor blocks into the topmost $L \left(L < N\right)$ layers for instruction-following fine-tuning. This can better tune the language representations with higher-level semantics. Taking the $l$-th inserted layer as an example, we denote the $M$-length word tokens as $T_{l} \in \mathbb{R}^{M \times C}$, where $C$ equals the feature dimension of LLaMA's attention layers. As shown in Figure \ref{fig:Overview}, the Excitor block takes $T_{l}$ with a set of learnable prompts $\{ P_{l} \}_{l=1}^{L}$ as inputs, where $P_{l} \in \mathbb{R}^{K \times C}$ with $K$ denoting the length of learnable prompts for each layer, generates an extra similarity matrix $S_{l}^{\rm{extra}} \in \mathbb{R}^{M \times M}$ with exactly the same shape of original similarity matrix $S_{l}$ for the $l$-th layer. In the self-attention mechanism, $S_{l}$ is calculated from $Query$ and $Key$, and controls the importance of components in $Value$, where $Query$, $Key$, and $Value$ are linear projected version of $T_{l}$:
\begin{small}
\begin{equation}
\begin{aligned}
Query = W_{\rm{q}} \left( T_{l} \right),
Key =   W_{\rm{k}} \left( T_{l} \right),
Value = W_{\rm{v}} \left( T_{l} \right)
\end{aligned}
\label{equation1}
\end{equation}
\end{small}
\noindent
where $W_{\rm{q}}$, $W_{\rm{k}}$, $W_{\rm{v}}$ are the pre-trained LLMs' weights that are frozen during the finetuning. Then, $S_{l}$ is formulated as:
\begin{small}
\begin{equation}
    S_{l} = \frac{Query \cdot Key^{\rm{T}}}{\sqrt{C}},
    \label{equation4}
\end{equation}
\end{small}
Since $S_{l}$ rules how the information updates, we treat $S_{l}^{extra}$ as the residual of $S_{l}$ to modify the reasoning process and formulate the updated output of the $l$-th layer $T_{l}^{\rm{o}}$ as:
\begin{small}
\begin{equation}
    S_{l}^{g} = {\rm{Softmax}} \left( S_{l}^{\rm{extra}} \times g_{l} + S_{l} \right),
    \label{equation5}
\end{equation}
\begin{equation}
    T_{l}^{\rm{o}} = W_{\rm{o}} \left( S_{l}^{g} \cdot Value \right),
    \label{equation6}
\end{equation}
\noindent
\end{small}
where $g_{l}$ is a set of learnable parameters, named cold-start gating factor, which is inspired by zero-init attention \cite{zhang2023llama}, to stabilize training by controlling the proportion of $S_{l}^{extra}$ participants in the reasoning. Instead of applying zero-initialization like \cite{zhang2023llama}, we initialize $g_{l}$ with $\mathbf{N}\left(0, 10^{-2}\right)$ to overcome the gradient vanishing in the mixed precision (FP16) training.

As stated in sec \ref{sec:intro}, previous techniques updating the output of frozen LLMs' layers by adding the outcomes of trainable modules onto $T_{l}^{\rm{o}}$ (e.g., Adapter\cite{houlsby2019parameter} and LoRA\cite{hu2021lora}) or concatenating trainable tokens with $T_{l}$ as input for layer $l$ (e.g., Prefix-tuning \cite{li2021prefix}) can be attributed to the direct modifying of the intermediate representations. One drawback of direct modifying is adding 
out-of-distribution features into the intermediate representations. It will largely change the inputs of higher layers and make the framework unstable, especially at the beginning of training. By contrast, LLaMA-Excitor only uses trainable tokens to change the attention score $S_{l}^{g}$. The output of each layer $T_{l}^{\rm{o}}$ is still the linear combination of $Value$ and within the original LLMs' distribution. i.e., LLaMA-Excitor alters the likelihood of potential outcomes generated by pre-trained LLMs. This indirect feature interaction ensures the model doesn’t deviate too far from its pretrained reasoning process and can more effectively arouse pretrained skills in downstream usage.

\subsection{Excitor Blocks}
\label{Excitor}
To impose effective control over the attention score $S_{l}^{g}$ while reducing the degradation of LLM's inherent abilities, LLaMA-Excitor must be self-adaptive to the original input sequence $T_{l}$ to decide whether learnable parameters should influence the reasoning. Thus, Excitor Blocks needs to generate the extra similarity matrix $S_{l}^{\rm{extra}}$ considering both injected information in the learnable prompt $P_{l}$ and $T_{l}$. We formulate $S_{l}^{\rm{extra}}$ as:
\begin{small}
\begin{equation}
    S_{l}^{\rm{extra}} = \frac{Query \cdot Key_{extra}^{\rm{T}}}{\sqrt{C}},
    \label{equation7}
\end{equation}
\noindent
\end{small}
where $Query$ is reused from the equation \ref{equation1} to pass knowledge from $T_{l}$ while reducing the computational overhead, $Key_{extra}$ is an extra attention key that carries information from $P_{l}$. Since the shape of $S_{l}^{\rm{extra}}$ should correspond with $S_{l}$, the shape of $Key_{extra}$ must be the same with $Key \in \mathbb{R}^{M \times C}$ in the equation \ref{equation1}. However, $M$, which is the length of $Key$, is changing along with variable input texts, while the length of learnable prompts $P_{l}$ is a pre-defined hyper-parameter. 

As shown on the left of the Excitor Block in Figure \ref{fig:Block Details}, we introduce the $Key$ reconstruction to mitigate the shape inconsistency. Since the attention mechanism generates an output with the same shape as its queries and transmits information about its values, we define $P_{l}$ as the Exitor $Key$ and the Exitor $Value$ and calculate the Excitor $Query$ from $T_{l}$ for the $l$-th Excitor Block. We omit linear projections for $P_{l}$ to pass information directly to $Key_{extra}$ and accelerate the training. The $Key$ reconstruction is formulated as:
\begin{small}
\begin{equation}
    S_{l}^{\rm{excitor}} = \frac{W_{\rm{q}}^{\rm{excitor}} \left( T_{l} \right) \cdot P_{l}^{\rm{T}}}{\sqrt{C}},
    \label{equation8}
\end{equation}
\begin{equation}
    Key_{\rm{extra}} = {\rm{Softmax}} \left( S_{l}^{\rm{excitor}} \right) \cdot P_{l},
    \label{equation9}
\end{equation}
\noindent
\end{small}
Where $W_{\rm{q}}^{\rm{excitor}}$ is a low-rank module consisting of two linear layers with dimension $\mathbb{R}^{C \times r}$ and $\mathbb{R}^{r \times C}$. Finally, $S_{l}^{\rm{extra}}$ carries information from $T_{l}$ and has the same shape as $Key$.

\begin{table*}[!t]
\fontsize{6pt}{0pt}\selectfont{
\begin{minipage}[t]{0.4\textwidth} 
\fcolorbox{black}{white!10}{\parbox{1\textwidth}
{
{\fontsize{8pt}{18pt}\selectfont
\textbf{Instruction: } \color{blue}{Tell me about alpacas.}}

\vspace{0.15cm}
\textbf{LLaMA7B: }Alpacas are a type of camelid that are native to South America. They are related to llamas and vicunas. Alpacas are raised for their fiber, which is used to make clothing.

\vspace{0.15cm}
\textbf{Full Fine-Tuning: }Alpacas are small, soft-furry, and mammals that are native to South America. They are the largest of the wool-making mammals and are the most gentle and social of the mammals. They are also the most colorful mammal, with a range of colors from white to gray, brown, black, and shades of brown and gray. They are also very smart and can learn tricks easily. They are also very social and live in herds of up to 20-30 individuals.

\vspace{0.15cm}
\textbf{Alpaca-LoRA: }Alpacas are a species of South American camelid, related to the llama and the vicuña. They are a domesticated version of the wild vicuña and are primarily raised for their fine fleece, which is used to make clothing. Alpacas are typically smaller than llamas, with a range of sizes from 12 to 15 hands high. They are known for their gentle nature and are often kept as pets.

\vspace{0.15cm}
\textbf{LLaMA-Adapter: }Alpacas are members of the camelid family and are native to the Andes Mountains of South America. They are typically between 3 and 4 feet tall and weigh between 100 and 200 pounds. They have long, coarse, and dense coats that come in a variety of colors, including white, black, brown, and grey. 

\vspace{0.15cm}
{\color{red}\textbf{LLaMA-Excitor:}} Alpacas are domesticated animals that are related to llamas and are native to South America. They are typically between 100 and 120 cm tall and weigh between 100 and 150 kg. Alpacas have long, soft fleece that is used for making clothing, blankets, and other products. They are also used for their meat and their dung, which is used as fuel. Alpacas are herd animals and are known for their gentle and calm nature. They are also very intelligent and can be trained to perform certain tasks.
}
 }
\end{minipage}
 \hspace{0.5cm}
\begin{minipage}[t]{0.55\textwidth} 
\fcolorbox{black}{white!10}{\parbox{1\textwidth}
{
 {\fontsize{8pt}{18pt}\selectfont
  \textbf{Instruction: } \color{blue}{Write a conversation between the sun and pluto.}}

\vspace{0.15cm}

\setlength\columnsep{0.0cm}
    \begin{minipage}[t]{0.3\textwidth}  
    \textbf{LLaMA7B: }
    
     The sun: Hey Pluto, how are you?

    \vspace{0.1cm}
     Pluto: I'm fine.

    \vspace{0.1cm}
     The sun: I'm glad to hear that.

    \vspace{0.1cm}
     Pluto: I'm glad to hear that too.

    \vspace{0.1cm}
     The sun: I'm glad to hear that too.
    \end{minipage}
    \hspace{0.2cm}
    \begin{minipage}[t]{0.3\textwidth}  
    {\textbf{Full Fine-Tuning: }
    
    The sun said, "I'm so hot!"

    \vspace{0.1cm}
    The sun said, "I'm so hot!"

    \vspace{0.1cm}
    The sun said, "I'm so hot!"

    \vspace{0.1cm}
    The sun said, "I'm so hot!"

    \vspace{0.1cm}
    The sun said, "I'm so hot!"

    }
    \end{minipage}
    \hspace{-0.3cm}
    \begin{minipage}[t]{0.4\textwidth}  
    {\textbf{Alpaca-LoRA: }
    
    Sun: Hey Pluto, how's life in the outer solar system?

    \vspace{0.1cm}
    Pluto: It's pretty chilly up here! I'm still getting used to being so far away from the sun.

    \vspace{0.1cm}
    Sun: That's understandable. I'm sure you'll get used to it soon enough. Anyway, I'm glad you're still around.

    \vspace{0.1cm}
    Pluto: Thanks! It's nice to still be a part of the solar system.
    }
    \end{minipage}  
   
    \vspace{-0.2cm}
    \begin{minipage}[t]{0.4\textwidth}  
    {\textbf{LLaMA-Adapter: }
    
    The Sun: Good morning, Pluto!

    \vspace{0.1cm}
    Pluto: Good morning, Sun! How are you?

    \vspace{0.1cm}
    The Sun: I'm doing great! How are you?

    \vspace{0.1cm}
    Pluto: I'm doing well, too!

    \vspace{0.1cm}
    The Sun: I'm glad to hear that!

    \vspace{0.1cm}
    Pluto: Me too!

    \vspace{0.1cm}
    The Sun: I'm glad we're having such a nice day!

    \vspace{0.1cm}
    Pluto: Me too!

    \vspace{0.1cm}
    The Sun: I'm glad we're having such a nice day!

    \vspace{0.cm}
    Pluto: Me too!
    }
    \end{minipage}  
    \begin{minipage}[t]{0.55\textwidth}  
    {{\color{red}\textbf{LLaMA-Excitor: }}
    
    Sun: Hey Pluto, how are you doing?
    
    Pluto: I'm doing great! How about you?
    
    Sun: I'm doing well too. It's been a while since we last spoke.
    
    Pluto: Yeah, it has been a while. I've been busy exploring the outer reaches of the solar system.
    
    Sun: That's great! I'm glad you're having fun.
    
    Pluto: I'm having a blast! I'm so glad I'm not stuck in the inner solar system like the other planets.
    
    Sun: Yeah, it's nice to have some space to yourself.
    
    Pluto: Definitely! So, what have you been up to?
    
    Sun: I've been busy keeping the other planets warm and providing light and heat.
    
    Pluto: That's a big job! You must be exhausted.
    
    Sun: I am, but it's worth it. I'm glad I can help.
    
    Pluto: I'm glad you're doing well. It's been great talking to you.
    }
    \end{minipage} 
}
 }
\end{minipage}
}
\caption{Instruction-following Comparison between LLaMA-Excitor and other techniques.}
\label{table:instructions}
\end{table*}
\subsection{Multi-modal Extension}
\label{Multi}
\begin{figure}
    \centering
    \includegraphics[width=0.4\textwidth]{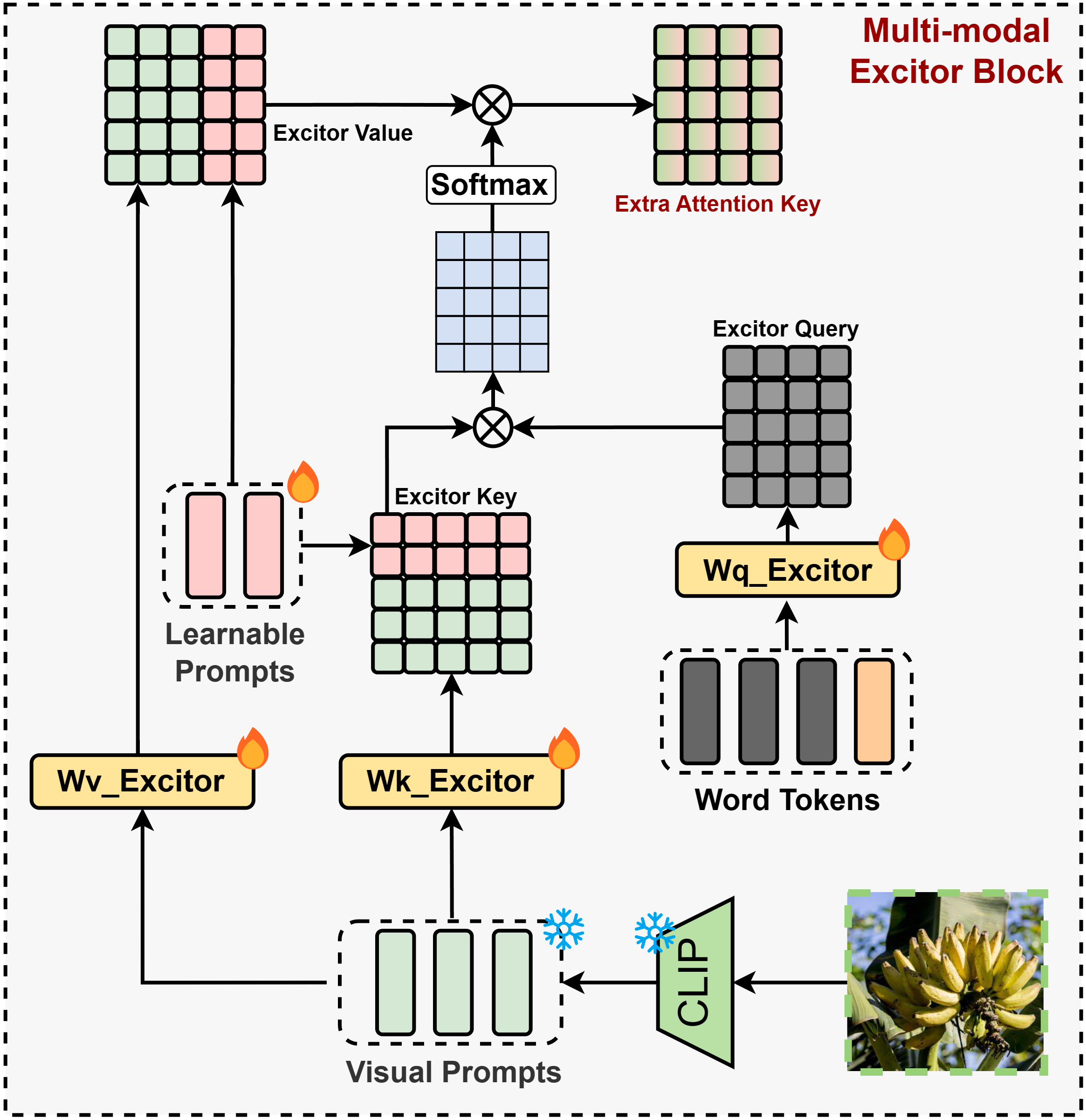}
    \caption{Extend LLaMA-Excitor into a powerful multi-modal model. Owing to the indirect feature interaction, LLaMA-Excitor is the cheapest PEFT method that can follow visual instructions without complicated projection modules aligning vision and language.}
    \label{fig:visual block}
\end{figure}

The indirect feature interaction is not limited to textual instructions but is also compatible with following visual prompts. We simply extend our Excitor Block into a multi-modal version in Figure \ref{fig:visual block}. Taking an image and a task description as input, LLaMA-Excitor first leverages a frozen pre-trained visual encoder like CLIP \cite{radford2021learning} to extract a sequence of visual embeddings $I \in \mathbb{R}^{V \times D}$ as the visual prompts, where $V$ is the embedding length and $D$ is the dimension of embeddings. Note that we adopt both the global CLS token and regional patch tokens from the last layer of the CLIP encoder to acquire adequate visual semantics. Then, we calculate multi-modal versions of the Excitor $Key$ and the Exitor $Value$ by combining $P_{l}$ and $I$:
\begin{small}
\begin{equation}
    Key_{\rm{excitor}} = \left[P_{l}; ~~ W_{\rm{k}}^{\rm{excitor}} \left( I \right)\right],
    \label{equation10}
\end{equation}    
\begin{equation}
    Value_{\rm{excitor}} = \left[P_{l}; ~~ W_{\rm{v}}^{\rm{excitor}} \left( I \right)\right].
    \label{equation11}
\end{equation}
\noindent
\end{small}
Unlike the processing of $P_{l}$, we adopt low-rank linear projections $W_{\rm{k}}^{\rm{excitor}}$ and $W_{\rm{v}}^{\rm{excitor}}$ on $I$, since we provide the same visual prompt for attention layers $l \in \left[ N-L, N\right]$ and different layers require various visual clues for the downstream reasoning.
\begin{figure}[t]
    \centering
    \includegraphics[width=0.45\textwidth]{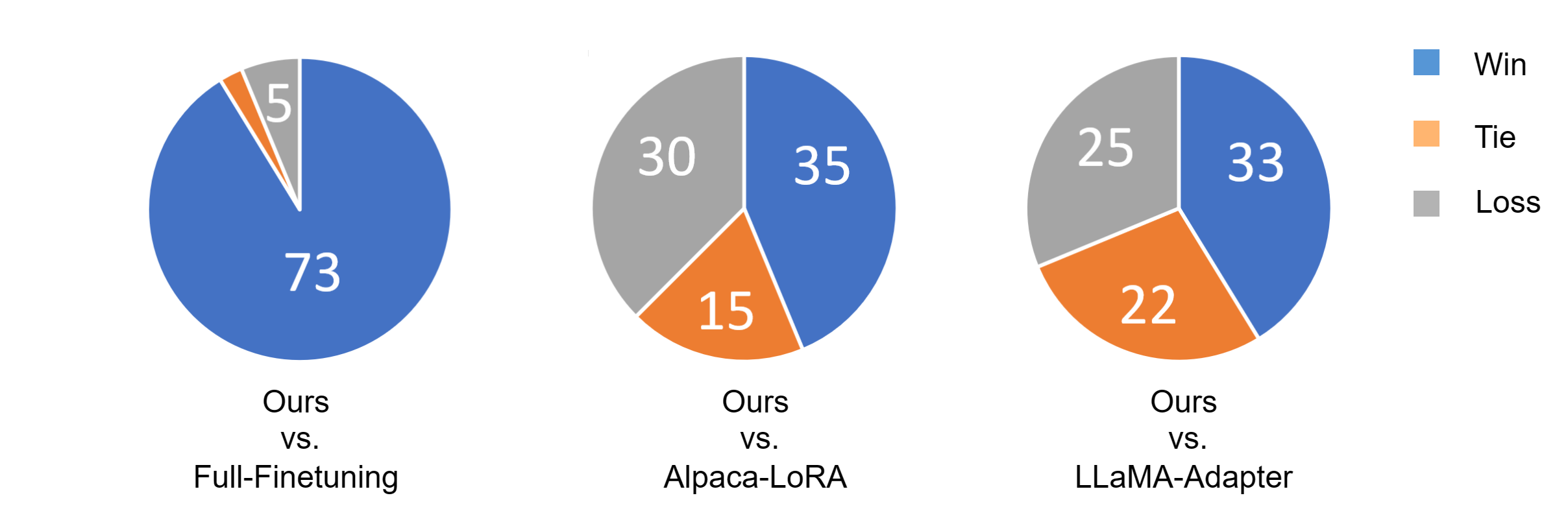}
    \caption{Quantitative comparisons between LLaMA-Excitor (BLUE) with other methods, evaluated by GPT-4 \cite{openai2023gpt}.}
    \label{fig:gpt4_vs}
\end{figure}
LLaMA-Excitor is an innovative framework for the multi-modal fine-tuning of LLMs. As mentioned in Section 1, most techniques before LLaMA-Excitor involved training a module to perform multi-modal alignment that projects visual embeddings into the textual feature space. This enables LLMs to understand visual tokens and integrate them into the reasoning process. However, the strict alignment constraint may damage the rich semantics of visual embeddings, and their direct feature fusion may also cause performance degradation. In contrast, LLaMA-Excitor does not require feature alignment and works on the raw inputs from the frozen image encoder. It does not change the purity of the intermediate textual features and thus avoids any potential damage to the rich semantics of the visual embeddings and the reasoning of LLMs.

\section{Experiments}
\label{sec:exp}
\subsection{Language-only Performances}
\subsubsection{Instruction-following Evaluations}
\label{Instruction-following Evaluations}
In this section, we evaluate the instruction-following capacity of LLaMA-Excitor by responding to instructions.

\noindent
\textbf{Dataset.} We use Stanford Alpaca \cite{taori2023stanford}, which contains 52K instruction-following data, as the training set. Each sample in this dataset is formatted according to a template as follows: 

\noindent
\fcolorbox{black}{white!10}{\parbox{0.95\linewidth}{
\small
\noindent
Below is an instruction that describes a task, paired with an input that provides further context. Write a response that appropriately completes the request. 

Instruction: \{instruction\} 

Input: \{input\} 

Response: \{output\}
}}
\noindent
where \{instruction\} is the description of a task, \{input\} is the context for the task, and \{output\} is the answer generated by GPT-3.5 \cite{brown2020language}.

\noindent
\textbf{Implementation Details.} We develop LLaMA-Excitor based on the original LLaMA codebase with minor modifications. We train LLaMA-Excitor on 8 NVIDIA A100 GPUs for five epochs. The warmup epochs, batch size, learning rate, and weight decay are set to 2, 64, 9e-3, and 0.02, respectively. In this section, we utilize the pretrained LLaMA-7B v1 model, which has $N=32$ transformer layers with feature dimension $C=4096$ as the backbone. We apply our excitor blocks with the low-rank dimension $r = 16$ for $W_{\rm{q}}^{\rm{excitor}}$ in the topmost $L=30$ layers and set the length of learnable prompts $L=30$. During the inference, we adopt top-p sampling as the default decoding method with a temperature of 0.1 and a top-p = 0.75.

We compare LLaMA-Excitor with three competing methods. (1) Full-finetuning, which updates the entire 7B parameters in LLaMA during the end-to-end training. (2) Alpaca-LoRA, the official implementation of low-rank adaptation \cite{hu2021lora} provided by \cite{taori2023stanford}. (3) LLaMA-Adapter \cite{zhang2023llama}, a cutting-edge prefix-tuning method with zero-init attention to reduce forgetting in finetuning.
\begin{figure}[t]
    \centering
    \includegraphics[width=0.5\textwidth]{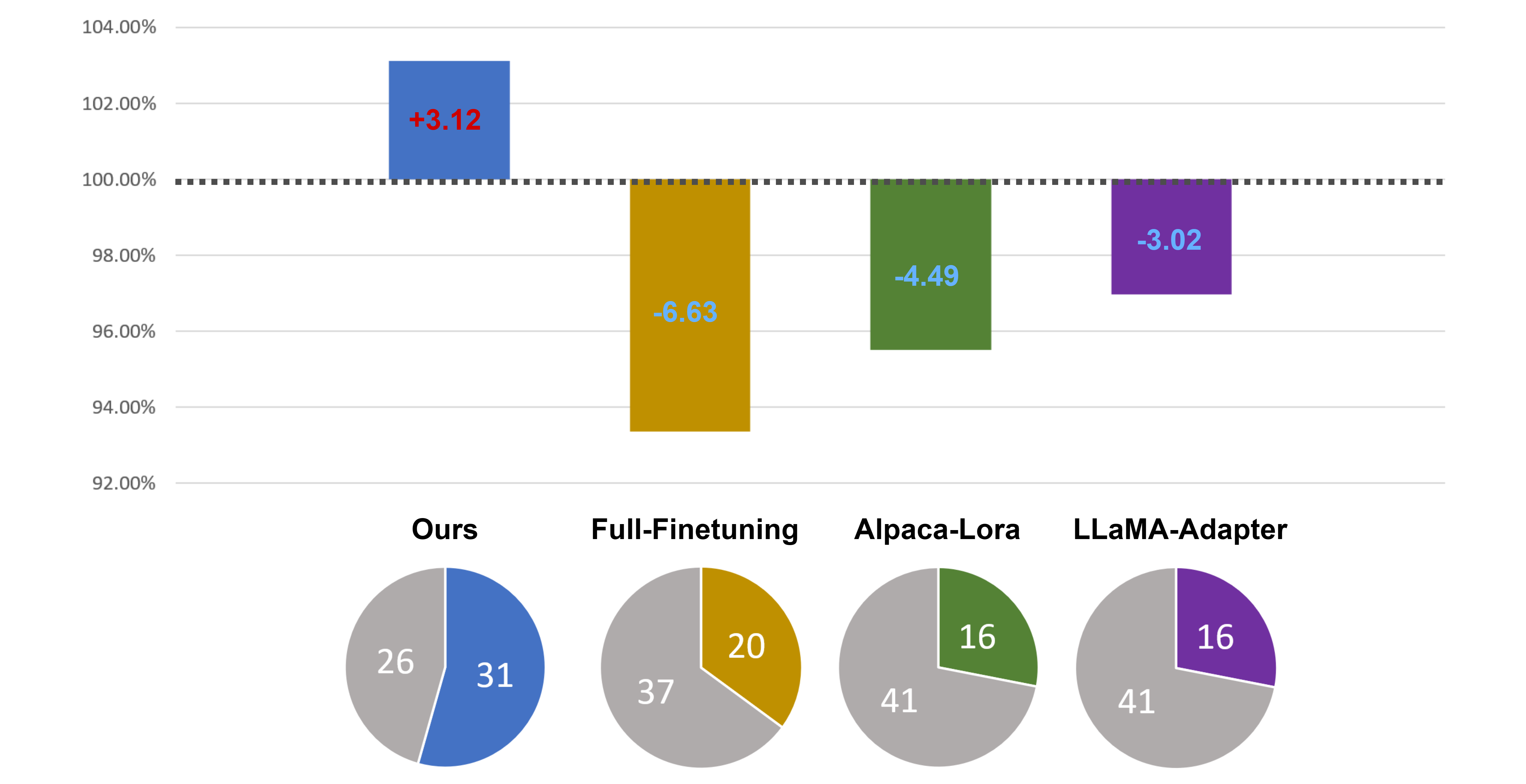}
    \caption{The relative average performance changes and win-loss situations of fine-tunings compared to original LLaMA-7B on MMLU. }
    \label{fig:mmlu_vs}
\end{figure}

\noindent
\textbf{Performances.} We first use examples in Table \ref{table:instructions} to show the strong instruction-following ability of LLaMA-Excitor. The first case requires the models to provide some information about alpacas. LLaMA-Excitor generates reasonable responses comparable to Full-finetuning, demonstrating our method's effectiveness. It's worth noticing that our approach follows the same answering pattern as the prefix-tuning method LLaMA-adapter and the original LLaMA7B (although Excitor provides more details). This phenomenon is reasonable since prefix-tuning methods use additional learnable prompts to guide the generation process of LLMs. They do not change the internal reasoning process, just like our Excitor. This characteristic may also be the foundation of avoiding catastrophic forgetting. For the second task, the LLMs are asked to create a dialogue between the Sun and Pluto. Out of all the models that are tested, only Excitor and LoRA correctly recognize the characters as planets within the Solar System and incorporate their identities into the conversation. However, Excitor stands out as it can generate lengthy, high-quality conversations rich in interesting content. (Please see Appendix A for the full comparison including dialog generation, code generation, and question answering.)

To evaluate the performance of instruction-following models quantitatively, we refer to the research conducted by Chiang et al. \cite{chiang2023vicuna}. In this evaluation, we use GPT-4 \cite{openai2023gpt} to assess the quality of responses to 80 randomly selected questions from AlpacaEval \cite{alpaca_eval}. However, we notice a bias in GPT-4's scoring, where it tends to give higher scores to the first response. To overcome this, we interchange the position of two responses and use the average score for the final rating. Results are shown in Figure \ref{fig:gpt4_vs}. We notice that Excitor overwhelms the Full-Finetuning (73/80). The reason behind this is that AlpacaEval is an out-of-distributed (OOD) test for our training set, and full-finetuning may easily overfit the training set. 
When compared with Alpaca-LoRA and LLaMA-Adapter, our Excitor still gains the majority of "Win", demonstrating our indirect feature interaction is at least comparable with or better than predomint PEFT methods. (Please refer to Appendix B for instances on AlpacaEval.)

As this paper is a preliminary attempt to study indirect feature interaction in LLMs and the 52K instruction-following data used in our study is automatically generated, it is possible to further enhance the performance of LLaMA-Excitor by improving the quality of the training data, adjusting the hyper-parameters to more reasonable values or exploring larger LLaMA models.
\begin{table*}
\centering
\tiny
\fontsize{8pt}{0pt}\selectfont{
\fcolorbox{black}{gray!10}{\parbox{0.9\textwidth}
{ 
\begin{minipage}{0.65\textwidth} 
{\fontsize{8pt}{18pt}\selectfont
\textbf{Instruction: }Please answer me based on this image. \color{blue}{Generate a caption that details what happened in the image.}}

\vspace{0.25cm}

{\color{red}\textbf{LLaMA-Excitor: }}An airplane is flying through the sky, with its landing gear down, as it approaches for a landing at an airport. The airplane is dominating the scene, with its bright red and white colors clearly visible against the backdrop of the cloudy sky. The airplane is about to touch down on the runway, creating a moment of excitement and anticipation for the passengers and crew on board.

\vspace{0.25cm}

\textbf{Human Annotation: A large passenger airplane flying through the air.}
\end{minipage}
\hspace{1cm}
\begin{minipage}{0.3\textwidth} 
{
\includegraphics[width=0.62\textwidth]{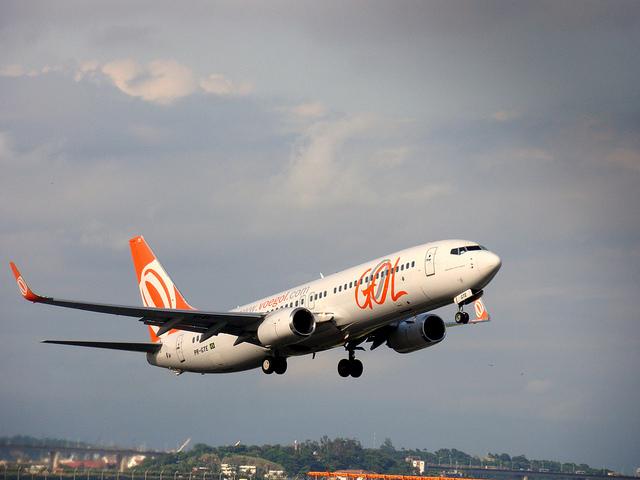}   
}
\end{minipage}
}
}
\fcolorbox{black}{gray!10}{\parbox{0.9\textwidth}
{ 
\begin{minipage}{0.65\textwidth} 
{\fontsize{8pt}{18pt}\selectfont
\textbf{Instruction: }Please answer me based on this image. \color{blue}{Generate a caption that details what happened in the image.}}

\vspace{0.25cm}

{\color{red}\textbf{LLaMA-Excitor: }}In the image, a group of birds is standing together on a girraffe. One of the bird is leaning over to nibble on the girraffe's fur and the giraffe seems to be enjoying the interaction. The scene captures the playful nature of these animals and the unexpected interaction between a giraffe and a bird.

\vspace{0.25cm}

\textbf{Human Annotation: Little birds sitting on the shoulder of a giraffe.}
\end{minipage}
\hspace{1cm}
\begin{minipage}{0.3\textwidth} 
{
\includegraphics[width=0.62\textwidth]{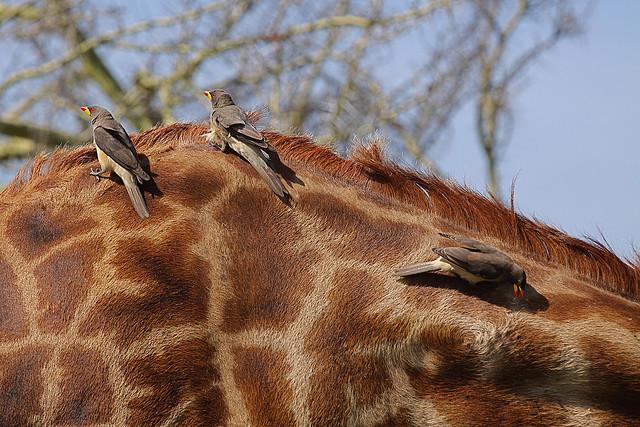}   
}
\end{minipage}
}
}

\vspace{0.1cm}






}
\caption{Examples demonstrating LLaMA-Excitor's visual instruction following capacity.}
\label{table:coco}
\vspace{-0.5cm}
\end{table*}
\subsubsection{The Impact of Fine-tuning Techniques on Inherent Abilities}
One of the motivations for introducing the indirect feature interaction in LLaMA-Excitor is to reduce the influence of fine-tuning on low-quality or non-targeted datasets. To quantitatively analyze this characteristic, we evaluate the fine-tuned models on MMLU \cite{hendrycks2020measuring}, which is currently the most comprehensive benchmark for assessing the multi-task accuracy of text models. This benchmark includes 57 tasks across various fields, such as basic math, American history, computer science, and law. The model must possess extensive world knowledge and problem-solving capabilities to achieve high accuracy on this test. We report the relative performance changes of PEFT methods compared with the original LLaMA-7B and how many of the 57 evaluations PEFT methods outperform LLaMA-7B in Figure \ref{fig:mmlu_vs}. We find that for models fine-tuned on Alpaca-52k, only Excitor can avoid reducing the average score of 57 tasks in MMLU, and it even shows an increase in performance (\inc{+3.12\%}). As expected, the full-finetuning performance suffers the most severe decline as its parameters are most significantly changed (\dec{-6.63\%}), while the LLaMA-Adapter, which is based on prefix-tuning, experiences the least decline (\dec{-3.02\%}) besides Exictor. 
(We provide results based on LLaMA2-7B in Appendix C)
\subsection{Multi-Modal Performances}
\begin{table}[ht]
\centering
\resizebox{0.8\linewidth}{!}{
\begin{tabular}{c|cc|cc}
\toprule
\multirow{2}{*}{Method} & \multicolumn{2}{c}{Data Scale} & \multicolumn{2}{c}{COCO Caption} \\
& PT & FT & BLEU@4 & CIDEr \\
\midrule
ClipCap \cite{mokady2021clipcap}        & 0M     & 0.6M  & 33.5  & 113.1\\
VL-PET \cite{hu2023vl}                  & 0M     & 0.6M  & -     & 121.7\\
LLaMA-AdapterV2 \cite{gao2023llama}     & 0M     & 0.6M  & 36.2  & 122.2\\
Qwen-vl-chat \cite{bai2023qwen}         & 1.4B   & 0.6M  & -     &131.9 \\
mPLUG-Owl2 \cite{ye2023mplug}           & 348M   & 0.6M  & -     & 137.3\\
BLIP \cite{li2022blip}                  & 14M    & 0.6M  & 40.4  & 136.7\\
Flamingo \cite{alayrac2022flamingo}     & 1.8B   & 0.6M  & -     & 138.1\\
BLIP-2 \cite{li2023blip}                & 129M   & 0.6M  & 43.7  & 145.3\\
\midrule
LLaMA-Excitor                           & 0M     & 0.6M  & \textbf{49.7}  & \textbf{157.5}\\
\bottomrule
\end{tabular}
}
\captionof{table}{Comparison with state-of-the-art image captioning methods on COCO Caption.}
\vspace{-0.5cm}
\label{tab:karpathy}
\end{table}
In this section, we evaluate the visual instruction following the capacity of LLaMA-Excitor by responding to paired vision-language instructions and demonstrate our indirect feature interaction, which unifies the modeling of language-only tuning and multi-modal tuning, providing a powerful low-budget way to perform vision-language tasks.
\subsubsection{Image Captioning Evaluation}
\textbf{Implementation Details.} We first evaluate our Excitor on COCO Caption \cite{chen2015microsoft}, which contains 0.6M training image caption data (120k images with 5 captions per image) over a wide range of distributions. We apply frozen CLIP/14-L \cite{radford2021learning} as the image encoder, the visual embedding dimension $D=768$, low-rank dimension $r=16$. We keep other hyper-parameters the same as the section \ref{Instruction-following Evaluations}.
\begin{table*}[ht]
    \centering
    \tiny
    \resizebox{0.9\textwidth}{!}{%
    \begin{tabular}{l|c|ccc|ccc|cc}
    \toprule
    \multirow{2}{*}{Model} & \multirow{2}{*}{Average}  &\multicolumn{3}{c|}{Subject} &\multicolumn{3}{c|}{Context Modality}   &\multicolumn{2}{c}{Grade}\\      
    
    & & NAT               & SOC               & LAN               & TXT               & IMG               & NO                & G1-6              & G7-12             \\ \midrule
    Human~\cite{saikh2022scienceqa}                                             & 88.40            & 90.23             & 84.97             & 87.48             & 89.60             & 87.50             & 88.10             & 91.59             & 82.42              \\ \midrule
    UnifiedQA$_{CoT}$                                                           & 74.11            & 71.00             & 76.04             & 78.91             & 66.42             & 66.53             & 81.81             & 77.06             & 68.82              \\
    GPT-3$_{CoT}$                                                               & 75.17            & 75.44             & 70.87             & 78.09             & 74.68             & 67.43             & 79.93             & 78.23             & 69.68              \\ 
    ChatGPT$_{CoT}$~\cite{chatgpt}                                              & 78.31            & 78.82             & 70.98             & 83.18             & 77.37             & 67.92             & 86.13             & 80.72             & 74.03              \\
    GPT-4$_{CoT}$~\cite{openai2023gpt}                                          & 83.99            & 85.48             & 72.44             & 90.27             & 82.65             & 71.49             & 92.89             & 86.66             & 79.04             \\
    MM-COT \cite{zhang2023multicot}                                             & 84.91            & 87.52             & 77.17             & 85.82             & 87.88             & 82.90             & 86.83             & 84.65             & 85.37              \\ 
    LLaVA$_{CoT}$ \cite{liu2023visual}  & 90.92 & 90.36 & 95.95 & 88.00 & 89.49 & 88.00 & 90.66 & 90.93 & 90.90\\
    LLaVA$_{CoT}$ (w/o pretrain) \cite{liu2023visual} & 85.81 & - & - & - & - & - & - & - & -\\
    \midrule
    DFAF \cite{gao2019dynamic}                                                  & 60.72            & 64.03             & 48.82             & 63.55             & 65.88             & 54.49             & 64.11             & 57.12             & 67.17              \\
    ViLT \cite{pmlr-v139-kim21k}                                                & 61.14            & 60.48             & 63.89             & 60.27             & 63.20             & 61.38             & 57.00             & 60.72             & 61.90              \\
    Patch-TRM \cite{lu2021iconqa}                                               & 61.42            & 65.19             & 46.79             & 65.55             & 66.96             & 55.28             & 64.95             & 58.04             & 67.50              \\
    VisualBERT \cite{li2019visualbert,li2020does}                               & 61.87            & 59.33             & 69.18             & 61.18             & 62.71             & 62.17             & 58.54             & 62.96             & 59.92              \\
    UnifiedQA \cite{khashabi2020unifiedqa}                                      & 70.12            & 68.16             & 69.18             & 74.91             & 63.78             & 61.38             & 77.84             & 72.98             & 65.00              \\
    GPT-3 \cite{brown2020language}                                              & 74.04            & 75.04             & 66.59             & 78.00             & 74.24             & 65.74             & 79.58             & 76.36             & 69.87              \\
    LLaMA-Adapter & 85.19   & 84.37    & 88.30    & 84.36    & 83.72    & 80.32   & 86.90    & 85.83    & 84.05\\
    LLaMA-Excitor                  & 85.41 & 85.70 & \textbf{92.35} & 82.82 &83.43 & 84.56 & 86.27 & 85.65 & 84.64 \\
    LLaMA-Excitor@336px $+$ LoRA  & \textbf{88.39} &\textbf{ 87.19} & 91.33 & \textbf{87.09} &\textbf{90.42} &\textbf{ 85.20} & \textbf{88.64} & \textbf{88.35} & \textbf{88.42} \\
    \bottomrule
    
    \end{tabular}%
    }
    \caption{Question Answering Accuracy (\%) on ScienceQA's~\cite{saikh2022scienceqa} test set.
     We report GPT-3~\cite{brown2020language}, ChatGPT~\cite{chatgpt}, and GPT-4~\cite{openai2023gpt} for zero-shot inference. $CoT$ denotes utilizing chain-of-thought for question answering.}
    \label{tab:scienceqa}
\end{table*}

\noindent
\textbf{Performances.} We compare our Excitor with cutting-edge image captioning methods in Table \ref{tab:karpathy}. We find that Excitor achieves a stunning result, significantly surpassing the previous SOTA method BILP-2 \cite{li2023blip} by \inc{+6} BLEU@4 and \inc{+12.2} CIDEr. Especially considering that we do not apply complex multi-modal alignment modules and have only 0.6M training data. By contrast, BLIP-2 adopts a costly pre-training stage on an additional 129M dataset, including VisualGenome \cite{krishna2017visual}, Conceptual Captions \cite{sharma2018conceptual,schuhmann2021laion} and LAION \cite{changpinyo2021conceptual}. We visualize the results of image captioning in Appendix D. However, the image captions provided in COCO are concise and lack sufficient detail in describing the content of images. To further unleash Exitor's performance in image captioning, we continue fine-tuning Exitor on LLaVA665k \cite{liu2023improved}, a higher-quality dataset with detailed picture descriptions. We provide several examples from the COCO test set in Table \ref{table:coco}. It shows that image captions generated by Excitor can accurately cover the content of human annotations and provide richer details. (e.g., (1) The airplane's landing gear is down, (2) A bird is pecking at the back of a giraffe)

\subsubsection{Performances on ScienceQA}
\textbf{Implementation Details.} We evaluate our Excitor on ScienceQA \cite{saikh2022scienceqa}, which contains 21k multimodal multiple choice questions with rich domain diversity across 3 subjects, 26 topics, 127 categories, and 379 skills. We train Excitor on ScienceQA train split from scratch, using the combination of Chain-of-Thought(CoT) and direct answer prediction. CoT requires the model to predict the solution first and then generate the answer choice based on the solution. During training, we randomly require the model to first generate solutions or directly predict answers in each iteration. The model is asked to answer the question directly in the evaluation.

Table \ref{tab:scienceqa} reports the performances of the top methods on the official leaderboard of the ScienceQA and Excitor. 
The current SOTA method LLaVA \cite{liu2023visual} is pretrained on another 558k visual-language dataset and fine-tuned on ScienceQA. Also, they allow all LLM's parameters to be updated during fine-tuning. Our Excitor demonstrates comparable performance with LLaVA w/o pretraining (only \dec{-0.4\%} performance gap). Note that LLaMA-Excitor is a PEFT method (with a frozen LLM) without CoT, and LLaVA utilizes a larger backbone LLaMA-13B than our LLaMA-7B. To further unlock the potential of Excitor, we introduce a variant that utilizes a more powerful image encoder (CLIP-L/14@336) and adds LoRA blocks to the original LLM's parameters to reduce the gap with LLaVA's full parameter updating. It brings \inc{+2.58\%} performance gain compared with LLaVA.

\subsection{Ablations}
\begin{table}[h]
\centering
\tiny
\resizebox{0.8\linewidth}{!}{
\begin{tabular}{l| c}
\toprule
Variant                                 &ACC\\
\midrule
(a) CLIP-16/B                               &83.11\\
(b) CLIP-14/L                               &85.41\\
(c) CLIP-14/L@336px                         &85.87\\
(d) CLIP-14/L@336px + Full-Finetuning       &83.20\\
(e) CLIP-14/L@336px + LoRA                  &\textbf{88.39}\\
\bottomrule
\end{tabular}
}
\caption{Analysis of the effectiveness of each module.}
\label{table:effectiveness}
\end{table}
We ablate several design choices on ScienceQA in Table \ref{table:effectiveness}. (a), (b) and (c) study the influence of the capacity of image encoder. The best CLIP encoder yields 85.41\% and is \inc{+2.76\%} higher than the basic version. In (d) and (e), we study to combine Excitor, which focuses on instruction-following, with other techniques that update LLM's reasoning process. We find LoRA can bring \inc{2.98\%} performance gain for Excitor, demonstrating the potential of combining Excitor with existing finetuning methods in multi-modal usage. (However, in the text-only scenario, we encounter \dec{-4.14\%} performance gap compared with the original LLaMA-7B in the MMLU test by combining Excitor and LoRA.) Please see Appendix E for more ablations on the design of low-rank linear layers (i.e., $W_{\rm{q}}^{Excitor}$, $W_{\rm{k}}^{Excitor}$, and $W_{\rm{v}}^{Excitor}$) in Excitor Blocks under single-modal and multi-modal scenarios, the choice of low-rank dimension $r$, and the number of layers $L$ with Excitor blocks inserted.
%

\section{Conclusion}
In this paper, we study the indirect feature interaction in fine-tuning LLMs into instruction-following models and propose LLaMA-Exictor, a PEFT method that demonstrates outstanding instruction-following capacity compared with predominant techniques meanwhile reducing the forgetting of LLMs' inherent abilities. Moreover, Excitor also shows its potential in vision-language tasks by unifying the modeling of visual instruction-following and language-only instruction-following. Impressively, Excitor achieves strong multi-modal performance without the need to train complex alignment modules.

This project is a work in progress, and some directions can be explored. The first one is the adaptability to LLMs other than LLaMA. The second direction is to explore more effective ways to introduce multi-scale visual prompts (since we only apply the features from the last layer of CLIP at the current stage). Last but not least, Excitor has the potential to fine-tune vision-only models into vision-language models, just like our implementation for LLMs.

\section*{Acknowledge}
This work is supported in part by the Beijing Natural Science Foundation (No. L222024), and the National Natural Science Foundation of China (No. 62394322). One of the authors, Chao Yang, is supported by the Shanghai Post-doctoral Excellent Program (Grant No. 2022234).

\clearpage
{
    \small
    \bibliographystyle{ieeenat_fullname}
    \bibliography{main}
}
\clearpage
\appendix
\onecolumn
\cleardoublepage

\section{More Instruction-following Comparisons for Finetuning Techniques}
\begin{table*}[h]
\fontsize{8pt}{0pt}\selectfont{
\begin{minipage}[t]{1\textwidth} 
\fcolorbox{black}{white!10}{\parbox{1\textwidth}
{
{\fontsize{8pt}{18pt}\selectfont
\textbf{Instruction: } \color{blue}{Write a Python program that prints the first 10 Fibonacci numbers.}}

\vspace{0.15cm}
\textbf{LLaMA7B: }

\fcolorbox{black}{gray!10}{\parbox{0.95\textwidth}
{
fibonacci(10)

[1, 1, 2, 3, 5, 8, 13, 21, 34, 55]
}}
Explanation:

The fibonacci function takes a number as an argument and returns the next number in the Fibonacci sequence.

\vspace{0.15cm}
\textbf{Full Fine-Tuning: }

\fcolorbox{black}{gray!10}{\parbox{0.95\textwidth}
{
def print\_fibonacci\_10(n):

~~~~~~if n == 0:

~~~~~~~~~~~~print("0")

~~~~~~elif n == 1:
    
~~~~~~~~~~~~print("1")

~~~~~~else:
    
~~~~~~~~~~~~print("1" + str(Fibonacci(n - 1)) + "1")
}}

\vspace{0.15cm}
\textbf{Alpaca-LoRA: }

\fcolorbox{black}{gray!10}{\parbox{0.95\textwidth}
{
def fibonacci(n):

~~~~~~if n == 0:

~~~~~~~~~~~~return 0

~~~~~~elif n == 1:

~~~~~~~~~~~~return 1

~~~~~~else:
    
~~~~~~~~~~~~return fibonacci(n - 1) + fibonacci(n - 2)

print(fibonacci(10))
}}

\vspace{0.15cm}
\textbf{LLaMA-Adapter: }

\fcolorbox{black}{gray!10}{\parbox{0.95\textwidth}
{
def fibonacci(): 

~~~~~~a=0 

~~~~~~b=1 

~~~~~~for i in range(10): 

~~~~~~~~~~~~a, b = b, a + b 

~~~~~~~~~~~~print(a, b) 

~~~~~~print("Fibonacci numbers: ", a, b) 

fibonacci()
}}

\vspace{0.15cm}
{\color{red}\textbf{LLaMA-Excitor:}} 

\fcolorbox{black}{gray!10}{\parbox{0.95\textwidth}
{
def fibonacci(n):

~~~~~~if n == 0:

~~~~~~~~~~~~return 0

~~~~~~elif n == 1:

~~~~~~~~~~~~return 1

~~~~~~else:
    
~~~~~~~~~~~~return fibonacci(n - 1) + fibonacci(n - 2)

print(fibonacci(10))
}}

Output:

\fcolorbox{black}{gray!10}{\parbox{0.95\textwidth}{
[0, 1, 1, 2, 3, 5, 8, 13, 21, 34]
}}

Explanation:

The first line of the program defines a function named `fibonacci` that takes a single argument, `n`. The function returns the first `n` Fibonacci numbers.
}}
\end{minipage}
\vspace{0.65cm}
 
\begin{minipage}[t]{1\textwidth} 
\fcolorbox{black}{white!10}{\parbox{1\textwidth}
{
{\fontsize{8pt}{18pt}\selectfont
\textbf{Instruction: } \color{blue}{Write a shouting match between Julius Caesar and Napoleon.}}

\vspace{0.15cm}
\textbf{LLaMA7B: }

A shouting match between Julius Caesar and Napoleon.

\vspace{0.15cm}
\textbf{Full Fine-Tuning: }

Julius Caesar: "I am Caesar! I am the greatest!" 

Napoleon: "I am Napoleon! I am the greatest!"

\vspace{0.15cm}
\textbf{Alpaca-LoRA: }

Julius Caesar and Napoleon were having a heated argument. Julius was insisting that the Roman Empire was the greatest empire of its time and that its power should be respected. Napoleon, on the other hand, argued that the Roman Empire was outdated and that his own French Empire was far superior. Julius was incensed and demanded that Napoleon acknowledge his superiority. Napoleon, however, refused to back down and the shouting match continued.

\vspace{0.15cm}
\textbf{LLaMA-Adapter: }

Julius Caesar: ”You’re nothing but a petty tyrant, Napoleon! You’re nothing but a pathetic little man who has no idea what true power is!” 

Napoleon: ”You’re nothing but a weak and pathetic fool, Caesar! You’re nothing but a petty tyrant who has no idea what true power is!”

\vspace{0.15cm}
{\color{red}\textbf{LLaMA-Excitor:}} 

Julius Caesar: "I am the greatest general in the Roman Empire! I have conquered Gaul and I am now on my way to conquer Britain. I will be the greatest 

~~~~~~~~~~~~~~~~~~~~~~~~~emperor in history!"

Napoleon: "You are nothing but a common soldier! I am the greatest general in history! I have conquered Europe and I am now on my way to conquer the 

~~~~~~~~~~~~~~~~~~~world! I will be the greatest emperor in history!"
}}
\end{minipage}
 \hspace{0.5cm}

}
\end{table*}

\clearpage

\begin{table*}[h]
\fontsize{8pt}{0pt}\selectfont{
 
\begin{minipage}[t]{1\textwidth} 
\fcolorbox{black}{white!10}{\parbox{1\textwidth}
{
{\fontsize{8pt}{18pt}\selectfont
\textbf{Instruction: } \color{blue}{List all Canadian provinces in alphabetical order.}}

\vspace{0.15cm}
\textbf{LLaMA7B: }

Alberta, British Columbia, Manitoba, New Brunswick, Newfoundland and Labrador, Nova Scotia, Ontario, Prince Edward Island, Quebec, Saskatchewan, Yukon

\vspace{0.15cm}
\textbf{Full Fine-Tuning: }

Alberta, British Columbia, Manitoba, New Brunswick, Newfoundland and Labrador, Nova Scotia, Ontario, Prince Edward Island, Quebec, Saskatchewan, Northwest Territories, Nunavut, Yukon.

\vspace{0.15cm}
\textbf{Alpaca-LoRA: }

Alberta, British Columbia, Manitoba, New Brunswick, Newfoundland and Labrador, Nova Scotia, Ontario, Prince Edward Island, Quebec, Saskatchewan.

\vspace{0.15cm}
\textbf{LLaMA-Adapter: }

Alberta, British Columbia, Manitoba, New Brunswick, Newfoundland and Labrador, Northwest Territories, Nova Scotia, Nunavut, Ontario, Prince Edward Island, Quebec, Saskatchewan, Yukon.

\vspace{0.15cm}
{\color{red}\textbf{LLaMA-Excitor:}} 

1. Alberta 

2. British Columbia 

3. Manitoba 

4. New Brunswick 

5. Newfoundland and Labrador 

6. Northwest Territories 

7. Nova Scotia 

8. Nunavut 

9. Ontario 

10. Prince Edward Island 

11. Quebec 

12. Saskatchewan 

13. Yukon Territory
}}
\end{minipage}
}
\end{table*}
\section{Examples of LLaMA-Excitor's responses on Alpaca-Eval}
\begin{table*}[h]
\fontsize{8pt}{0pt}\selectfont{
 
\begin{minipage}[t]{1\textwidth} 
\fcolorbox{black}{white!10}{\parbox{1\textwidth}
{
{\fontsize{8pt}{18pt}\selectfont
\textbf{Instruction: } \color{blue}{Hi, I'm trying to solve a crossword puzzle, but I've never done one of these before. Can you help me out?}}

\vspace{0.15cm}
{\color{red}\textbf{LLaMA-Excitor:}} 

Sure! Here are some tips to get you started: 

1. Look for clues that are repeated in the puzzle. These are usually the most important clues.

2. Look for words that are in the same category. For example, if you see the word "cat" in one clue, 

~~~~look for words that are similar to it, such as "kitten" or "pussycat".

3. Look for words that are antonyms or synonyms of each other. For example, if you see the word "happy" in one clue, 

~~~~look for words that are opposite or similar to it, such as "sad" or "cheerful".

4. Look for words that are in the same word family. For example, if you see the word "dog" in one clue, 

~~~~look for words that are related to it, such as "bark" or "tail".

5. Look for words that are in the same part of speech. For example, if you see the word "run" in one clue, 

~~~~look for words that are verbs, such as "walk" or "jump".

6. Look for words that are in the same grammatical form. For example, if you see the word "the" in one clue, 

~~~~look for words that are articles, such as "a" or "an".
}}
\end{minipage}

\vspace{0.15cm}

\begin{minipage}[t]{1\textwidth} 
\fcolorbox{black}{white!10}{\parbox{1\textwidth}
{
{\fontsize{8pt}{18pt}\selectfont
\textbf{Instruction: } \color{blue}{I like to host guests at my home from time to time, and I am gathering recipes of different dishes and drinks to keep things interesting. I am interested in trying some Georgian dishes. Can you give me a recipe for Kubdari?}}

\vspace{0.15cm}
{\color{red}\textbf{LLaMA-Excitor:}} 

Kubdari is a traditional Georgian dish made with ground meat, onions, and spices. It is usually served with a side of tomato sauce and is a popular dish in Georgia. Here is a recipe for Kubdari:

1. In a large skillet, heat 1 tablespoon of oil over medium-high heat.

2. Add the ground meat and cook until it is no longer pink.

3. Add the onions and cook until they are translucent.

4. Add the garlic, cayenne pepper, and salt and pepper to taste.

5. Cook for 5 minutes, stirring occasionally.

6. Add the tomato sauce and cook for 5 minutes.

7. Serve hot with a side of tomato sauce.
}}
\end{minipage}

\vspace{0.15cm}

\begin{minipage}[t]{1\textwidth} 
\fcolorbox{black}{white!10}{\parbox{1\textwidth}
{
{\fontsize{8pt}{18pt}\selectfont
\textbf{Instruction: } \color{blue}{Can you help me figure out how to sew a button on a shirt?}}

\vspace{0.15cm}
{\color{red}\textbf{LLaMA-Excitor:}} 

Yes, here is a step-by-step guide to sewing a button on a shirt:

1. Place the button on the shirt, making sure it is centered and facing the right direction.

2. Stitch the button to the shirt using a needle and thread.

3. Trim the excess thread.

4. Press the button down to make sure it is securely attached to the shirt.
}}
\end{minipage}

\vspace{0.15cm}

\begin{minipage}[t]{1\textwidth} 
\fcolorbox{black}{white!10}{\parbox{1\textwidth}
{
{\fontsize{8pt}{18pt}\selectfont
\textbf{Instruction: } \color{blue}{How do I make apple juice?}}

\vspace{0.15cm}
{\color{red}\textbf{LLaMA-Excitor:}} 

To make apple juice, you will need to wash and core the apples, then cut them into small pieces. Place the pieces in a large pot and add water. Bring the pot to a boil and simmer for 10 minutes. Strain the juice and store in an airtight container in the refrigerator.
}}
\end{minipage}
}
\end{table*}

\begin{table*}[t]
\centering
\tiny
\resizebox{0.8\linewidth}{!}{
\begin{tabular}{l l ccc | cc}
\toprule
Method & LLM & Res. & PT & IT & VQA$^\text{v2}$ & GQA\\
\midrule
BLIP-2 \cite{li2023blip}                & Vicuna-13B & 224 & 129M & -    & 41.0    & 41   \\
InstructBLIP \cite{dai2023instructblip} & Vicuna-7B  & 224 & 129M & 1.2M & --      & 49.2 \\
InstructBLIP \cite{dai2023instructblip} & Vicuna-13B & 224 & 129M & 1.2M & --      & 49.5 \\
Shikra \cite{chen2023shikra}            & Vicuna-13B & 224 & 600K & 5.5M & 77.4$^*$& --   \\
IDEFICS-9B \cite{IDEFICS}               & LLaMA-7B   & 224 & 353M & 1M   & 50.9    & 38.4 \\
IDEFICS-80B \cite{IDEFICS}              & LLaMA-65B  & 224 & 353M & 1M   & 60.0    & 45.2 \\
Qwen-VL \cite{bai2023qwen}              & Qwen-7B    & 448 & 1.4B$^\dagger$ & 50M$^\dagger$ & 78.8$^*$ & 59.3$^*$ \\
Qwen-VL-Chat \cite{bai2023qwen}         & Qwen-7B    & 448 & 1.4B$^\dagger$ & 50M$^\dagger$ & 78.2$^*$ & 57.5$^*$ \\
LLaVA1.5 \cite{liu2023improved}         & Vicuna-7B  & 336 & 558K & 665K & 78.5$^*$ &62.0$^*$ \\
LLaVA1.5 \cite{liu2023improved}         & Vicuna-13B & 336 & 558K & 665K &\underline{80.0}$^*$ &\textbf{63.3}$^*$ \\
\midrule
LLaMA-Excitor                           & LLaMA2-7B  & 336 & -    & 665k &\textbf{83.6$^*$} &\underline{62.1$^*$} \\
\bottomrule
\end{tabular}
}
\caption{\textbf{Comparison with SoTA methods on VQA-v2~\cite{goyal2017making} and GQA~\cite{hudson2019gqa}.} Res, PT, IT indicate input image resolution, the number of samples in the pretraining, and the instruction tuning stage, respectively. $^*$The training images of the datasets are observed during training. $^\dagger$Includes in-house data that is not publicly accessible.}
\label{tab:results}
\end{table*}
\section{Supplemental Results on QA tasks and The Property of Not Forgetting (based on LLaMA2)}
\textbf{Comparision on VAQ-v2 and GQA.}
We further evaluate the generalization ability of our LLaMA-Excitor on commonly used VQA-v2~\cite{goyal2017making} and GQA~\cite{hudson2019gqa}. Unlike our competitors, we do not pretrain to align image and text embeddings. We solely finetune LLaMA-Excitor on LLaVA665k \cite{liu2023improved} visual instruction-tuning dataset. From Table \ref{tab:results}, LLaMA-Excitor outperforms the cutting-edge methods by \inc{+3.6\%} on VQA-v2 and provides a competitive result (-1.2\%) on GQA. Since VQA-v2 and GQA are lean to evaluate the model's visual encoding ability rather than its ability to reason based on text, these results actually demonstrate the strong and impressive visual instruction-following ability of LLaMA-Excitor. Besides, it is promising to improve the Excitor further by better extracting visual prompts (currently, we simply utilize the original outputs from the last layer of a CLIP encoder).

\begin{figure}[ht]
    \centering
    \includegraphics[width=0.9\textwidth]{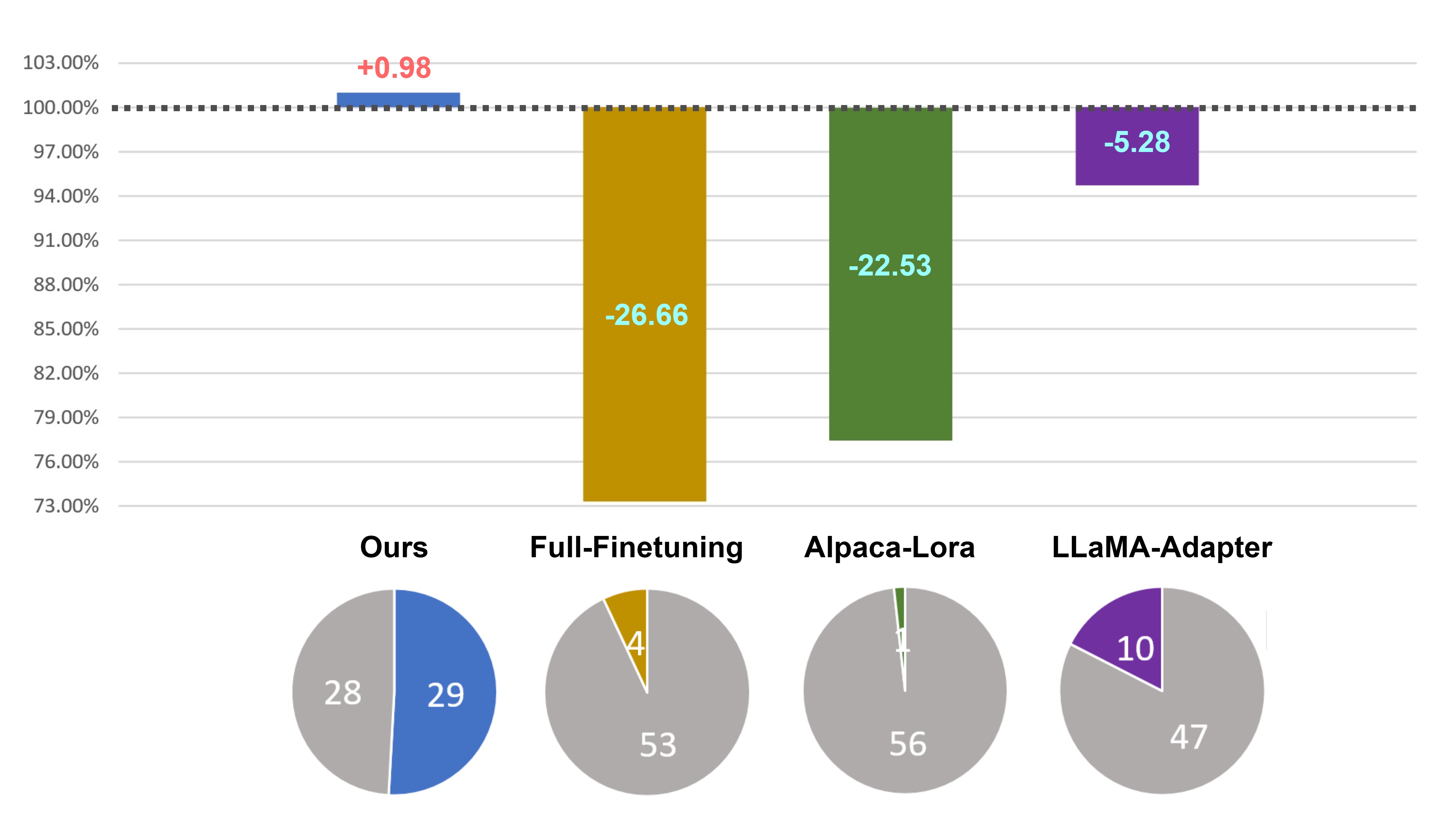}
    \caption{The relative performance changes and win-loss situations of fine-tunings compared to the original LLaMA2-7B on MMLU.}
    \label{fig:llama2_vs}
\end{figure}
\clearpage

\noindent
\textbf{The Impact of Fine-tuning to Inherent Abilities.}
We first conduct the same experiment as Sec. 4.1.2 for LLaMA2-7B and report the results in Figure \ref{fig:llama2_vs}. We find that as the model's ability improves, the performance degradation caused by fine-tuning on non-targeted datasets using previous methods has also significantly increased. LLaMA-Excitor maintains the overall MMLU performance (\inc{+0.98\%}) and mitigates catastrophic forgetting.

\begin{figure}[t]
    \centering
    \includegraphics[width=0.45\textwidth,height = 0.27\textwidth]{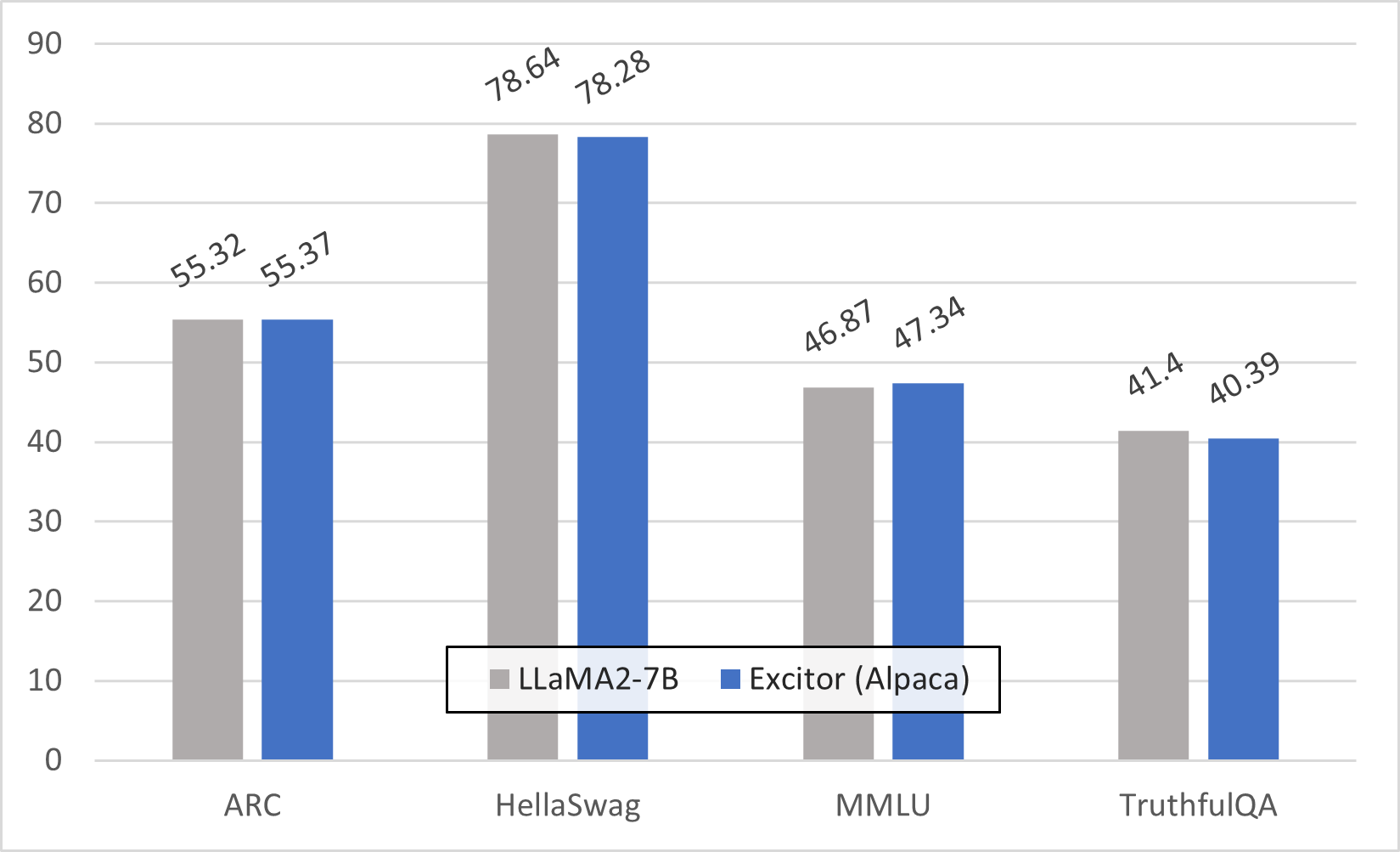}
    \includegraphics[width=0.45\textwidth,height = 0.27\textwidth]{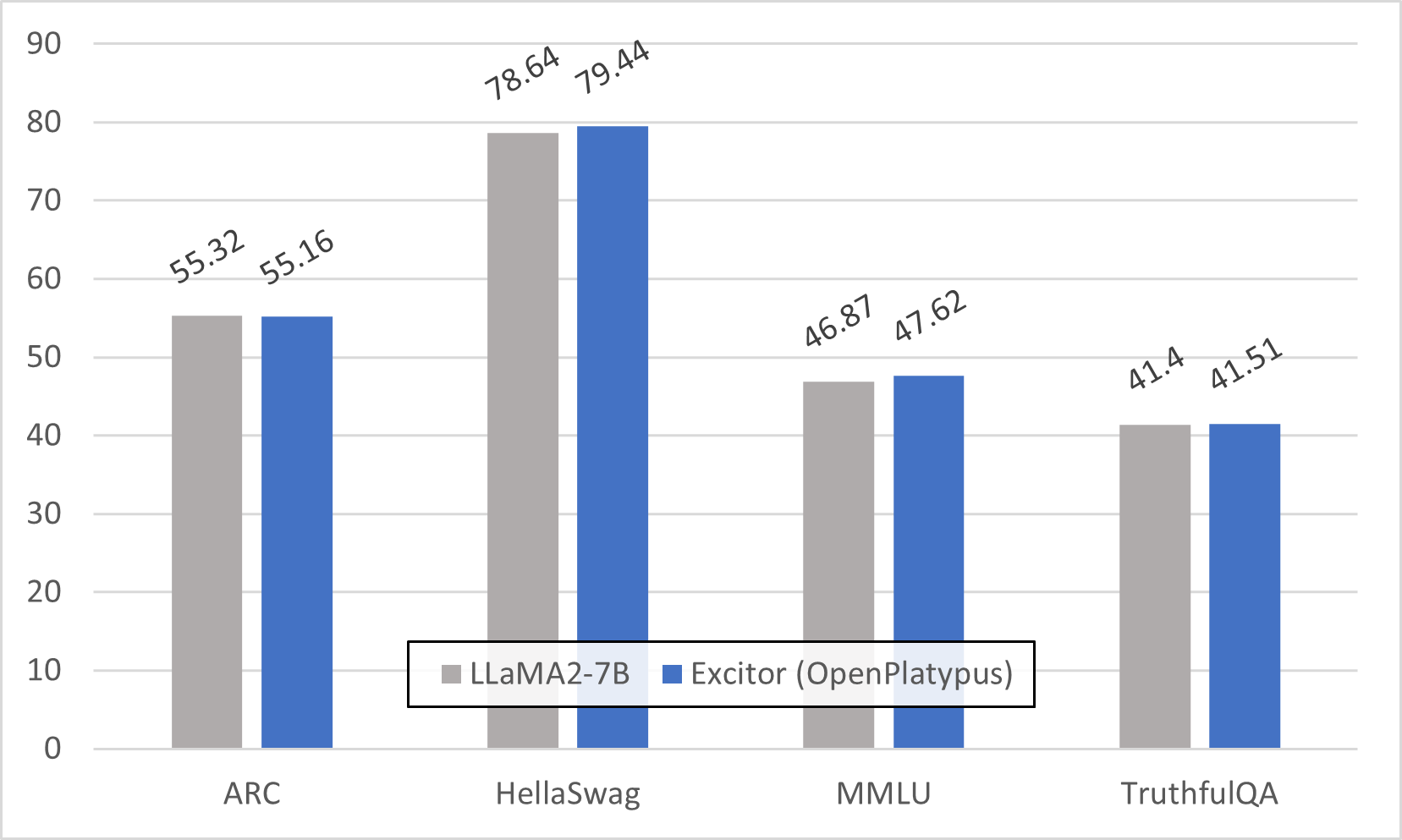}
    \par
    \includegraphics[width=0.45\textwidth,height = 0.27\textwidth]{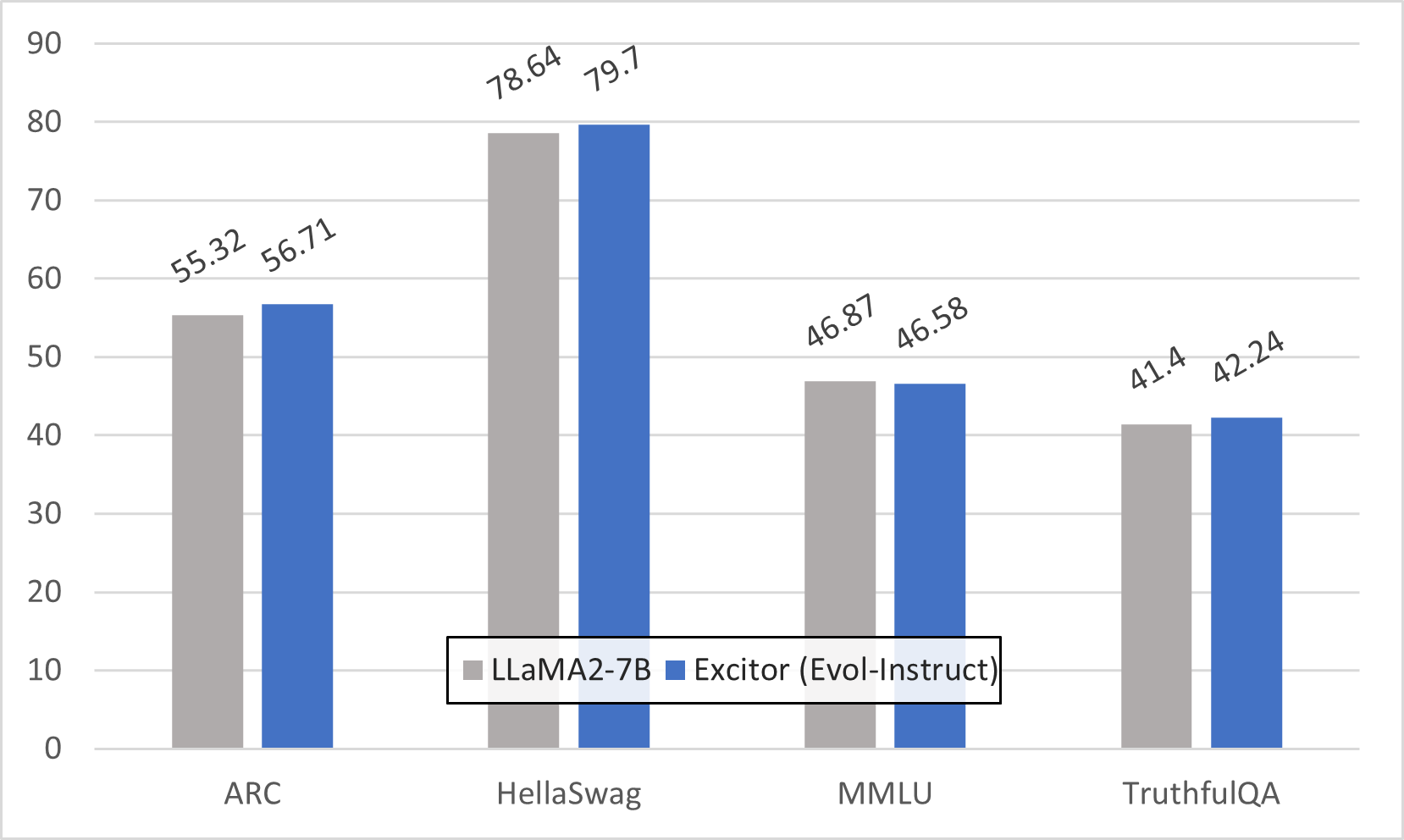}
    \caption{LLaMA-Excitor fine-tuned on Alpaca \cite{taori2023stanford}, OpenPlatypus \cite{lee2023platypus}, and Evol-Instruct \cite{xu2023wizardlm} compared with original LLaMA2-7B on ARC \cite{clark2018think}, HellaSwag \cite{zellers2019hellaswag}, MMLU \cite{hendrycks2020measuring}, and TruthfulQA \cite{lin2021truthfulqa}.}
    \label{fig:forgetting}
\end{figure}

To further verify the property of not forgetting, we fine-tune our LLaMA-Excitor on three instruction-tuning datasets (Alpaca \cite{taori2023stanford}, OpenPlatypus \cite{lee2023platypus}, and Evol-Instruct \cite{xu2023wizardlm}) and evaluate the performance change on four general benchmarks (ARC \cite{clark2018think}, HellaSwag \cite{zellers2019hellaswag}, MMLU \cite{hendrycks2020measuring}, and TruthfulQA \cite{lin2021truthfulqa}). As shown in Figure \ref{fig:forgetting}, no matter which finetuning dataset is used, Excitor's performance on the four tasks is always on par with LLaMA2-7B.

\clearpage

\section{Examples of Image Captioning on MSCOCO}
\begin{table*}[h]
\centering
\tiny
\fontsize{8pt}{0pt}\selectfont{

\fcolorbox{black}{white!10}{\parbox{0.8\textwidth}
{ 
\begin{minipage}{0.5\textwidth} 
{\fontsize{8pt}{18pt}\selectfont
\textbf{Instruction: }

Please answer me based on this image. \color{blue}{Generate a caption of the image.}}

\vspace{0.3cm}

{\color{red}\textbf{LLaMA-Excitor: }}

A clock made to look like a bike.

\vspace{0.3cm}

\textbf{Human Annotation:} 

A bicycle replica with a clock as the front wheel.
\end{minipage}
\hspace{2.5cm}
\begin{minipage}{0.3\textwidth} 
{
\includegraphics[width=0.5\textwidth]{figs/samples/0.jpg}   
}
\end{minipage}
}}

\fcolorbox{black}{white!10}{\parbox{0.8\textwidth}
{ 
\begin{minipage}{0.5\textwidth} 
{\fontsize{8pt}{18pt}\selectfont
\textbf{Instruction: }

Please answer me based on this image. \color{blue}{Generate a caption of the image.}}

\vspace{0.3cm}

{\color{red}\textbf{LLaMA-Excitor: }}

A bathroom sink with a variety of toiletries on it.

\vspace{0.3cm}

\textbf{Human Annotation:} 

A bathroom sink with toiletries on the counter.
\end{minipage}
\hspace{2.25cm}
\begin{minipage}{0.3\textwidth} 
{
\includegraphics[width=0.55\textwidth]{figs/samples/17.jpg}   
}
\end{minipage}
}}

\fcolorbox{black}{white!10}{\parbox{0.8\textwidth}
{ 
\begin{minipage}{0.5\textwidth} 
{\fontsize{8pt}{18pt}\selectfont
\textbf{Instruction: }

Please answer me based on this image. \color{blue}{Generate a caption of the image.}}

\vspace{0.3cm}

{\color{red}\textbf{LLaMA-Excitor: }}

A woman sitting on a bench next to a sidewalk.

\vspace{0.3cm}

\textbf{Human Annotation:} 

Two women waiting at a bench next to a street.
\end{minipage}
\hspace{2.25cm}
\begin{minipage}{0.3\textwidth} 
{
\includegraphics[width=0.55\textwidth]{figs/samples/19.jpg}   
}
\end{minipage}
}}

\fcolorbox{black}{white!10}{\parbox{0.8\textwidth}
{ 
\begin{minipage}{0.5\textwidth} 
{\fontsize{8pt}{18pt}\selectfont
\textbf{Instruction: }

Please answer me based on this image. \color{blue}{Generate a caption of the image.}}

\vspace{0.3cm}

{\color{red}\textbf{LLaMA-Excitor: }}

A box with a few black bananas inside of it.

\vspace{0.3cm}

\textbf{Human Annotation:} 

This is an open box containing four cucumbers.
\end{minipage}
\hspace{2.25cm}
\begin{minipage}{0.3\textwidth} 
{
\includegraphics[width=0.55\textwidth]{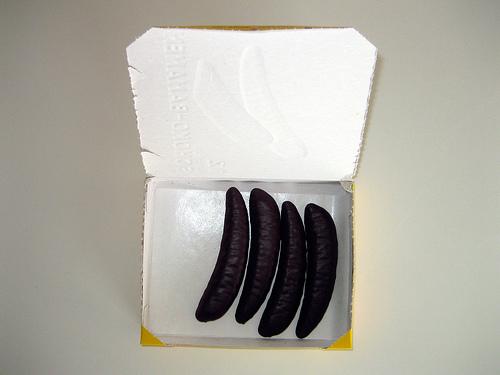}   
}
\end{minipage}
}}

\fcolorbox{black}{white!10}{\parbox{0.8\textwidth}
{ 
\begin{minipage}{0.5\textwidth} 
{\fontsize{8pt}{18pt}\selectfont
\textbf{Instruction: }

Please answer me based on this image. \color{blue}{Generate a caption of the image.}}

\vspace{0.3cm}

{\color{red}\textbf{LLaMA-Excitor: }}

A group of motorcyclists ride down a city street.

\vspace{0.3cm}

\textbf{Human Annotation:} 

Several motorcycles riding down the road in formation.
\end{minipage}
\hspace{2.25cm}
\begin{minipage}{0.3\textwidth} 
{
\includegraphics[width=0.55\textwidth]{figs/samples/27.jpg}   
}
\end{minipage}
}}

\fcolorbox{black}{white!10}{\parbox{0.8\textwidth}
{ 
\begin{minipage}{0.5\textwidth} 
{\fontsize{8pt}{18pt}\selectfont
\textbf{Instruction: }

Please answer me based on this image. \color{blue}{Generate a caption of the image.}}

\vspace{0.3cm}

{\color{red}\textbf{LLaMA-Excitor: }}

A cat is drinking water from a toilet.

\vspace{0.3cm}

\textbf{Human Annotation:} 

A black cat is inside a white toilet.
\end{minipage}
\hspace{2.25cm}
\begin{minipage}{0.3\textwidth} 
{
\includegraphics[width=0.55\textwidth]{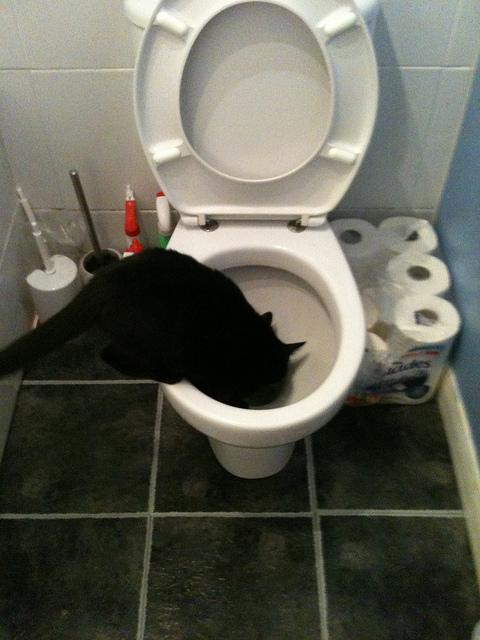}   
}
\end{minipage}
}}

\fcolorbox{black}{white!10}{\parbox{0.8\textwidth}
{ 
\begin{minipage}{0.5\textwidth} 
{\fontsize{8pt}{18pt}\selectfont
\textbf{Instruction: }

Please answer me based on this image. \color{blue}{Generate a caption of the image.}}

\vspace{0.3cm}

{\color{red}\textbf{LLaMA-Excitor: }}

A man sits with a decorated cow outside

\vspace{0.3cm}

\textbf{Human Annotation:} 

A man sits with a traditionally decorated cow
\end{minipage}
\hspace{2.25cm}
\begin{minipage}{0.3\textwidth} 
{
\includegraphics[width=0.55\textwidth]{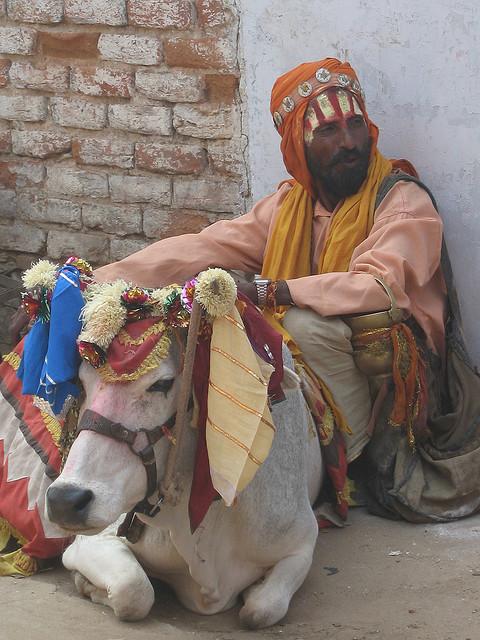}   
}
\end{minipage}
}}

}
\caption{Examples demonstrating LLaMA-Excitor's visual instruction following capacity.}
\label{table:coco_case}
\vspace{-0.5cm}
\end{table*}
\clearpage

\begin{table}[h]
\centering
\tiny
\resizebox{0.45\linewidth}{!}{
\begin{tabular}{l| ccc|c}
\toprule
\multirow{2}{*}{Variant}         &\multicolumn{3}{c}{Linear projection on $P_{l}$ and $T_{l}$}             &MMLU\\
                                 &$Query$         &$Key$            &$Value$                         &mACC(\%)                     \\
\midrule
(a)                              &\ding{55}     &\ding{55}      &\ding{55}                     &31.07\\
(b)                              &\checkmark    &\checkmark     &\checkmark                    &35.10\\
(c)                              &\ding{55}     &\ding{55}      &\checkmark                    &32.82\\
(d)                              &\ding{55}     &\checkmark     &\ding{55}                     &33.48\\
(e)                              &\checkmark    &\ding{55}      &\ding{55}                     &\textbf{35.75}\\
\bottomrule
\end{tabular}
}
\hspace{1cm}
\resizebox{0.4\linewidth}{!}{
\begin{tabular}{l| cc|c}
\toprule
\multirow{2}{*}{Variant}         &\multicolumn{2}{c}{Linear projection on $I$}                 &ScienceQA\\
                                 &$Key$          &$Value$                                       &mACC(\%)   \\
\midrule
(a)                              &\ding{55}      &\ding{55}                                    &75.96\\
(b)                              &\checkmark     &\ding{55}                                    &79.83\\
(c)                              &\ding{55}      &\checkmark                                   &81.20\\
(d)                              &\checkmark     &\checkmark                                   &\textbf{85.41}\\
\bottomrule
\end{tabular}
}
\caption{Analysis of the position of adding projection layers. Left: projection layers for learnable prompts and word tokens in text-only fine-tuning. Right: projection layers for frozen visual prompts in multi-modal fine-tuning.}
\label{table:more_ablation}
\end{table}

\begin{table}[h]
\centering
\tiny
\resizebox{0.3\linewidth}{!}{
\begin{tabular}{c| c}
\toprule
Low-Rank dimension  & MMLU\\
 $r$                &mACC(\%) \\
\midrule
4           &29.14\\
8           &34.00\\
16          &\textbf{35.75}\\
32          &34.21\\
64          &35.12\\
\bottomrule
\end{tabular}
}
\hspace{3cm}
\resizebox{0.325\linewidth}{!}{
\begin{tabular}{c | c}
\toprule
Number of Layers inserted  & MMLU\\
 $L$                        &mACC(\%) \\
\midrule
24          &34.50\\
26          &34.95\\
28          &35.32\\
30          &\textbf{35.75}\\
32          &35.66\\
\bottomrule
\end{tabular}
}
\caption{Analysis of the Low-Rank dimension $r$ and the number of layers $L$ with Excitor blocks inserted.}
\label{table:more_ablation_r_l}
\end{table}
\section{More Ablations}
We first study the best position of adding linear projection layers for learnable prompts $P_{l}$ and word tokens $T_{l}$ in text-only instruction tuning. The Excitor block relies on $T_{l}$ to generate $Query$ and $P_{l}$ to generate $Key$ and $Value$. The way projection layers are added in the generation will significantly influence the final performance. From Table \ref{table:more_ablation} (left), we find applying additional processing on $T_{l}$ by linear projection to generate $Query$ and direct passing $P_{l}$ as $Key$ and $Value$ can provide the best performance on MMLU. This is reasonable since $P_{l}$ is trainable and can be adaptively adjusted, while $T_{l}$ is generated from a frozen layer of LLaMA. After determining the structure of the single-modal Excitor, we further study the best way to apply projections of visual prompt $I$ in the multi-modal extension. As we mentioned in Sec. 3.3, since we provide the same visual prompt for each attention layer, Excitor relies on the projection layers to extract different visual information. On Table \ref{table:more_ablation} (right), Variant (e) with projection layers for both $Key$ and $Value$ generation has the best performance.

We adopt low-rank projection layers like LoRA \cite{hu2021lora} for Excitor blocks to reduce the number of trainable parameters and accelerate the training. Table \ref{table:more_ablation_r_l} (left) reports performances under different low-rank dimensions $r$ on MMLU. Besides, Table \ref{table:more_ablation_r_l} (right) provides the performance changes under different choices of the number of layers $L$ with Excior blocks inserted.


\end{document}